\documentclass[10pt,twocolumn,letterpaper]{article}
\usepackage{authblk}
\usepackage{cvpr}              

\usepackage[dvipsnames]{xcolor}

\definecolor{cvprblue}{rgb}{0.21,0.49,0.74}
\usepackage[pagebackref,breaklinks,colorlinks,citecolor=cvprblue]{hyperref}

\def\Vec#1{{\boldsymbol{#1}}}
\def\Mat#1{{\boldsymbol{#1}}}
\usepackage{mathtools}
\usepackage{booktabs}
\usepackage{algorithm}
\usepackage{algpseudocode}
\usepackage{makecell}
\usepackage{subcaption}
\usepackage[dvipsnames]{xcolor}
\usepackage{float}

\usepackage{amsmath}

\DeclarePairedDelimiter\abs{\lvert}{\rvert}%
\DeclarePairedDelimiter\norm{\lVert}{\rVert}%
\def\paperID{13378} 
\def\confName{CVPR}
\def\confYear{2024}

\title{Zoom-shot: Fast and Efficient Unsupervised \underline{Z}er\underline{o}-Sh\underline{o}t Transfer of CLIP to Vision Encoders with \underline{M}ultimodal Loss}

\begin{document}

\author[1]{Jordan Shipard}
\author[1,2]{Arnold Wiliem}
\author[1]{Kien Nguyen Thanh}
\author[3]{Wei Xiang}
\author[1]{Clinton Fookes}

\affil[1]{\small Signal Processing, Artificial Intelligence and Vision Technologies (SAIVT), Queensland University of Technology, Australia}
\affil[2]{Sentient Vision Systems, Australia}
\affil[3]{School of Computing, Engineering and Mathematical Sciences, La Trobe University, Australia}
\affil[ ]{\textit {\tt \small \{minhtuan.tran@connect, jordan.shipard@hdr, c.fookes@\}qut.edu.au}, \textit{ \tt \small \{arnoldw, hermawanm\}@sentientvision.com}}

\maketitle
\begin{abstract}
The fusion of vision and language has brought about a transformative shift in computer vision through the emergence of Vision-Language Models (VLMs). However, the resource-intensive nature of existing VLMs poses a significant challenge. We need an accessible method for developing the next generation of VLMs. To address this issue, we propose Zoom-shot, a novel method for transferring the zero-shot capabilities of CLIP to any pre-trained vision encoder. We do this by exploiting the multimodal information (\ie text and image) present in the CLIP latent space through the use of specifically designed multimodal loss functions. These loss functions are (1) cycle-consistency loss and (2) our novel prompt-guided knowledge distillation loss (PG-KD). PG-KD combines the concept of knowledge distillation with CLIP's zero-shot classification, to capture the interactions between text and image features.
With our multimodal losses, we train a \textbf{linear mapping} between the CLIP latent space and the latent space of a pre-trained vision encoder, for only a \textbf{single epoch}. Furthermore, Zoom-shot is entirely unsupervised and is trained using \textbf{unpaired} data. We test the zero-shot capabilities of a range of vision encoders augmented as new VLMs, on coarse and fine-grained classification datasets, outperforming the previous state-of-the-art in this problem domain. In our ablations, we find Zoom-shot allows for a trade-off between data and compute during training; and our state-of-the-art results can be obtained by reducing training from 20\% to 1\% of the ImageNet training data with 20 epochs. All code and models are available on GitHub.
\end{abstract}

\vspace*{-2.5mm} 
\section{Introduction}

\begin{figure}
    \centering
    \includegraphics[trim={3mm 3mm 3mm 2mm}, width=0.9\linewidth, clip]{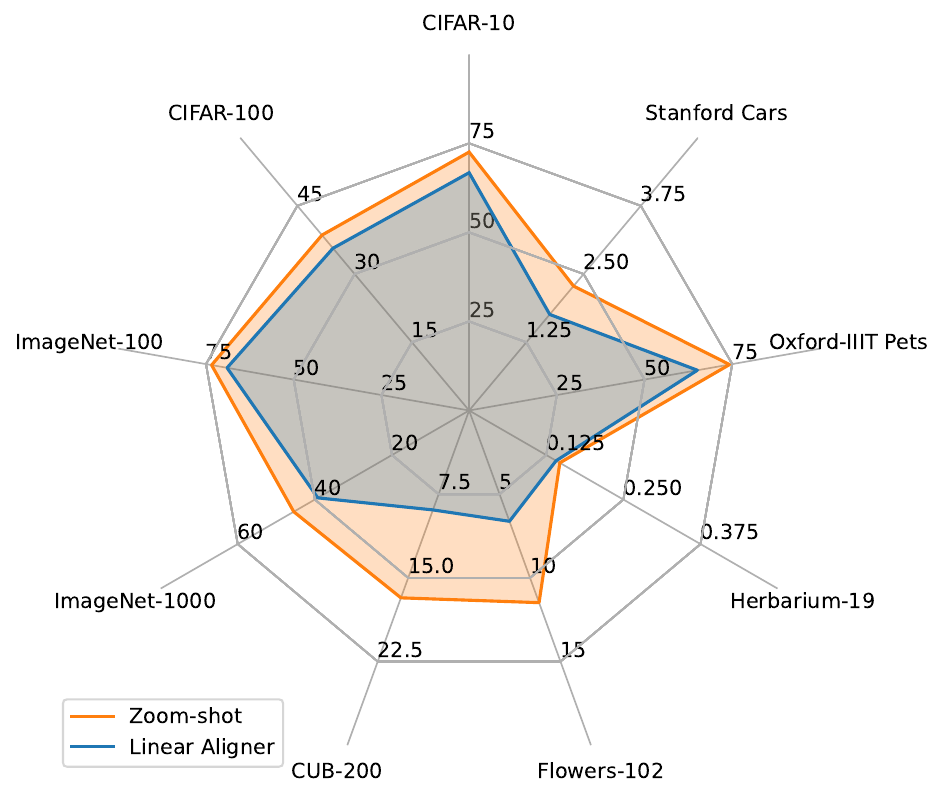}
    \caption{A summary of our results comparing the average top-1 zero-shot test accuracy of the recent state-of-the-art in this problem domain, Linear Aligner~\cite{t2c2023}, to our proposed Zoom-shot method. The averaged results are from MobileNetV3 small~\cite{Howard2019SearchingFM}, DenseNet-121~\cite{Huang2016DenselyCC}, ResNet-18~\cite{He2015DeepRL}, DINOv1 (ViT-B/16)~\cite{caron2021emerging} and DINOv2 (ViT-B/14)~\cite{dinov22023} vision encoders. We divide our testing datasets into coarse-grained (CIFAR-10/100~\cite{krizhevsky_learning_2009} and ImageNet-100/1000~\cite{deng_imagenet_2009}) and fine-grained (CUB-200~\cite{WahCUB_200_2011}, Flowers-102~\cite{Nilsback08}, Herbarium-19~\cite{tan2019herbarium}, Oxford-IIIT Pets~\cite{parkhi12a} and Stanford Cars~\cite{6755945}). Our method consistently outperforms the Linear Aligner.}
    \vspace{-4mm}
    \label{fig:summary}
\end{figure}
Computer vision has recently witnessed a paradigm shift with the integration of the vision and language domains leading to the creation of Vision-Language Models (VLMs). These VLMs, exemplified by CLIP~\cite{clip2021}, have enabled many applications, from zero-shot classification to image generation~\cite{SD2022, glide22, dalle2}. However, many of the existing VLMs demand significant resources for training and inference~\cite{flamingo2022, flava2022, albef2021, cyclip2022}. The cost of training especially limits the discovery of novel applications. It confines novel VLMs to being created by large organisations with access to significant computing resources. In many cases, combating this has required a huge effort by the open-source community~\cite{openclip}. Unfortunately, this all remains out of scope for the vast majority of research efforts. We need an accessible method, in terms of both compute and data, for developing the next generation of VLMs, such that they can be feasibly trained by researchers with access to limited computing resources.

With regard to this, we consider training from scratch to be infeasible. Instead, we consider methods that can transfer the knowledge inside existing VLMs into new ones. As an analogy, gaining the knowledge to write a book may take years, yet gaining the knowledge from reading it, may only take days.
A recent work, \cite{t2c2023}, demonstrates a promising route through cross-model alignment, obviating training from scratch by mapping between the latent spaces of pre-trained vision encoders. Surprisingly, this can be performed by learning \textit{only} a linear mapping. Doing so, one can augment a vision only model as a novel VLM via mapping to CLIP's joint vision-language latent space.
However, their method only focuses on mapping image features into a fundamentally multimodal latent space.
This is problematic, as the interaction between text and image features enables CLIP's VLM capabilities.
Moreover, the recently identified \textit{modality gap}~\cite{liang2022mind, shi2023towards} between CLIP's image and text features underscores that solely mapping image features only accounts for a subspace within CLIP's larger latent space. 

In our work, we propose Zoom-shot (\textbf{Z}er\textbf{O}-sh\textbf{O}t transfer with \textbf{M}ultimodal loss), named for its fast and efficient transfer of CLIP's zero-shot capabilities to arbitrary pre-trained vision encoders. Zoom-shot improves the quality of the learnt linear map by utilising multimodal loss functions. These loss functions are designed to capture the interaction between text and image features in CLIP's latent space. Furthermore, Zoom-shot is entirely unsupervised and is trained using unpaired data. We conduct zero-shot classification tests using five pre-trained vision encoders of varying size and capability (MobileNetV3 small~\cite{Howard2019SearchingFM}, DenseNet-121~\cite{Huang2016DenselyCC}, ResNet-18~\cite{He2015DeepRL}, DINOv1~\cite{dino2021}, DINOv2~\cite{dinov22023}). From the efficiency of MobileNetV3, to the robustness of DINOv2, these vision encoders were selected as they cover a range of capabilities. Additionally, it is important to examine our methods' performance across a range of image classification datasets. As such, we select three coarse grained and four fine-grained datasets of varying difficulty. The coarse grained datasets are CIFAR-10~\cite{krizhevsky_learning_2009}, CIFA-R100~\cite{krizhevsky_learning_2009} and ImageNet~\cite{deng_imagenet_2009}, with the fine-grained being CUB200~\cite{WahCUB_200_2011}, Flowers-102~\cite{Nilsback08}, Herbarium-19~\cite{tan2019herbarium}, Oxford-IIIT Pets~\cite{parkhi12a} and Stanford Cars~\cite{6755945}. 

Using our multimodal loss functions, Zoom-shot achieves a new state-of-the-art on nearly all tested datasets and models within this problem domain. A summary of our results is shown Figure \ref{fig:summary}. 
We conduct a number of ablation experiments to further investigate how the multimodal loss functions improve the quality of the linear mapping. Doing so, we find Zoom-shot posses a trade-off between compute and data; and show the linear mapping can be learnt using only 1\% of the ImageNet training data, given sufficient epochs. Additionally, we find the distribution of training images significantly impacts performance. We liken this approach to the vision encoders reading different books on different subjects (different training distributions), all written by the teacher, CLIP. We release all models and make our code available on GitHub.

\noindent Our main \textbf{contributions} are as follows:
\begin{enumerate}        
    \item We introduce multimodal loss functions for the unsupervised training of linear mapping functions between CLIP and pre-trained vision encoders, enhancing knowledge transfer in our Zoom-shot approach. Even though the training does not require paired image-text data, it still can capture important interactions between text and image features. 
    \item Our multimodal loss functions comprise two parts: the cycle-consistency loss~\cite{CycleGAN2017} and the novel prompt-guided knowledge distillation (PG-KD). Whilst the cycle-consistency loss has been widely used in the literature, our novel PK-KD is unique. By combining the concept of knowledge distillation~\cite{distill2014} with the CLIP zero-shot classification, PG-KD enables Zoom-shot to learn the interactions between text and image/vision.
    \item We demonstrate the importance of capturing these interactions by achieving state-of-the-art zero-shot performance over the previous method, \cite{t2c2023}. In comparison, \cite{t2c2023} only focus on mapping features from the vision encoder and require 6 epochs for training. Zoom-shot only requires a single epoch to achieve these impressive results.
    \item We conduct extensive ablation studies in order to fully understand our method. We discover Zoom-shot possesses a trade-off between compute and training data, meaning situations with limited data can be overcome with compensatory compute. Additionally, we find data coverage is a limiting factor.
\end{enumerate}


\begin{figure*}[t]
    \centering
    \includegraphics[width=\textwidth,trim={4mm 90mm 3mm 151mm},clip]{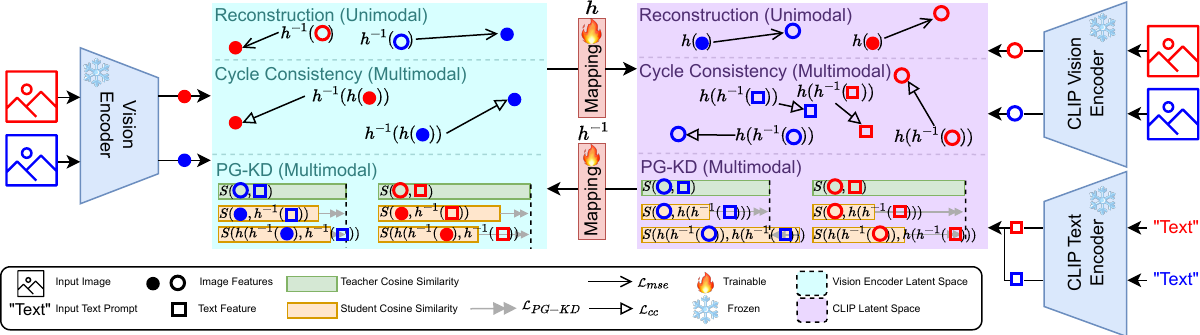}
    \caption{A diagram summarizing our Zoom-shot method. Zoom-shot trains a mapping function $h$ and its inverse $h^{-1}$ for mapping between a vision encoder's latent space and CLIP's latent space. The resulting mapping functions can be used to transfer the zero-shot performance of CLIP to an arbitrary vision encoder. At its core, Zoom-shot consists of three loss functions: reconstruction loss, cycle-consistency loss and Prompt-Guided Knowledge Distillation. We only display some of the student variants for PG-KD.}
    \vspace{-5mm}
    \label{fig:enter-label}
\end{figure*}
\vspace{-2mm}
\section{Related Work}
\vspace{-1mm}
\label{sec:related_works}

CLIP~\cite{clip2021} marked a milestone in vision-language models. Since its inception, various methods have surpassed its zero-shot accuracy through scaling data and compute~\cite{EVA-CLIP, openclip, zhai2023sigmoid, li2023clipa}. Other methods have focused on improving efficiency while maintaining performance~\cite{wang2021minivlm, kuang2023dlip, wang2022efficientvlm, derakhshani2023small}. Our approach, Zoom-shot, aims to address both ends of this spectrum, by retaining performance with both small (e.g., MobileNetV3~\cite{Howard2019SearchingFM}) and enhancing performance with larger vision encoders like DINOv2~\cite{dinov22023}. 

CLIP also demonstrated a path forward for multimodal models by establishing a joint latent space across modalities, which subsequent works expanded upon \cite{imageBind23, omnivore22, polyVit23}. This latent space has led to novel applications like zero-shot image generation \cite{SD2022, glide22, dalle2}. However, there are recently identified limitations, such as the modality gap present between CLIP's text and image features, as a result of local minimas in its training objective \cite{liang2022mind, shi2023towards}. Despite this, the CLIP's latent space may still contain untapped potential. Our focus involves transferring information from the CLIP latent space to pre-trained vision encoders, enabling them to leverage CLIP's zero-shot capabilities without retraining the encoders. 

Many different works have made progress on similar tasks, although with differing end goals to our own. For instance, LiT~\cite{lit22} aimed to improve the efficiency of training VLMs by contrastively fine-tuning a pre-trained text model to a pre-trained vision encoder. While an improvement over CLIP training from scratch, it still requires significant computational resources and data as it fine-tunes the whole encoder model. In a similar work, Merullo~\etal~ \cite{merullo2023linearly} align pre-trained vision encoders to pre-trained language models using only linear mapping. While this could hypothetically be used to create a CLIP-like VLM, they only use it to investigate the representations within language models (\ie mapping from image features into the text latent space). The work of Khan and Fu~\cite{contalign23} aims to address the shortcomings suffered by LiT by focusing on aligning only approximately 7\% of the model parameters. Although it reduces training parameters, it still requires large training data. In addition, all these works use contrastive learning which requires paired image and text data. Our method does not need paired data and requires a much smaller amount of data to train. This is due to leveraging the multimodal information encoded in the CLIP latent space.

Different from the above works, Moayeri~\etal~\cite{t2c2023} with their proposed Linear Aligner (LA), showed that a pre-trained vision encoder can be directly aligned to CLIP's latent space using only a linear mapping trained with unlabeled image data. To our knowledge, LA was the first method to show that it is possible to align a pre-trained vision encoder with CLIP's latent space, effectively augmenting the vision encoder with the VLM capabilities of CLIP, without the need for paired data.
Unfortunately, they only map the vision encoder, neglecting the language component and the fundamental interaction between these two modalities. Our approach, Zoom-shot, capitalizes on this interaction, leading to remarkable improvements in performance and training efficiency. 

\vspace{-2mm}
\section{Zoom-shot}
\vspace{-1mm}
\label{sec:method}
Overall, our aim is to transfer the zero-shot classification capabilities of CLIP~ to arbitrary vision encoders, without the use of labels. As shown in~\cite{t2c2023}, we can do this by augmenting a vision encoder through the use of a linear mapping function between the encoder's latent space and the CLIP latent space. The zero-shot, and broader vision-language capabilities of CLIP arise from its shared embedding space between its image and text encoder. Allowing for cosine similarity comparisons between features from each encoder, however, it was recently shown that each modality actually forms a separate subspace embedded within the latent space~\cite{liang2022mind, shi2023towards}. This separation of the subspaces is referred to as the \textit{modality gap}, and it exists due to local minimas in CLIP's contrastive training objective. We visually demonstrate the modality gap in the supplementary material.
This fundamentally changes how we approach the zero-shot transfer problem, as we need to account for these separate subspaces.

This section focuses on explaining our method, Zoom-shot, first by providing a technical problem formulation, which then leads into explaining our selected loss functions.


\def\targetvisenc{$V_t$}
\def\targettextenc{$T_t$}
\def\sourcevisenc{$V_s$}

\noindent
\textbf{Problem Formulation - }
We follow the problem formulation in~\cite{t2c2023} with slight modifications. Let \targetvisenc{}, \targettextenc{} be CLIP vision and text encoders, respectively. Let \sourcevisenc{} be the source vision encoder (\eg MobileNetV3~\cite{Howard2019SearchingFM}), $\Mat{x}$ be an input image, and $\Vec{p}$ be a tokenized prompt. \targettextenc{} maps $\Vec{p}$ into the text subspace embedded within the CLIP shared $d$-dimensional latent space, denoted as \targettextenc$(\Vec{p}) \in \mathbb{R}^d$. Similarly, \targetvisenc{} maps $\Mat{x}$, into the vision subspace embedded within the CLIP shared $d$-dimensional latent space: \targetvisenc$( \Mat{x} ) \in \mathbb{R}^d$. Our aim is to replace \targetvisenc{}  with \sourcevisenc{}, which maps $\Mat{x}$ into an $m$-dimensional latent space, \sourcevisenc{}$( \Mat{x}) \in \mathbb{R}^m$. As such, we train a linear function, $h$, to map from the $m$-dimensional latent space into the $d$-dimensional latent space, $h: \mathbb{R}^m \mapsto \mathbb{R}^d$. Note that, based on the assumption of a perfectly shared latent space, the original formulation in~\cite{t2c2023} confines $h$ to only transform the source vision latent space, into the CLIP vision latent space. In this work, we consider the source vision latent space to be a subspace embedded within the $m$-dimensional latent space. We call this latent space as the \textit{$m$-dimensional source latent space}. This definition allows us to define $h$ as a mapping from the $m$-dimensional source text/vision subspace into the $d$-dimensional target CLIP text/vision subspace\footnote{We do not consider cross-modal mapping (\eg vision-to-text)}. Additionally, we can extract relationship information between text and vision data used for training a better mapping function. Once $h$ is learned, we can then use it to transform features extracted by \sourcevisenc{} into the CLIP latent subspace.



\vspace{-1mm}
\subsection{Loss Functions}
The challenge in solving this problem is that we are only learning a linear mapping; therefore, we need to carefully consider our loss functions, as they have a significant impact on guiding the optimization. 
An overall diagram of our loss functions can be seen in Figure~\ref{fig:enter-label}.
The two key components proposed in~\cite{t2c2023} are, firstly, the use of reconstruction loss, and secondly, the importance of re-scaling the variance of the features within the two latent spaces. As previously stated, these components only focus on guiding the mapping function to learn the vision encoder subspace. Due to the modality gap, applying reconstruction loss may only help the mapping function learn the information from the CLIP vision subspace.
We argue the information from the text subspace, and the relationship between text and vision features, are crucial for training the better mapping function. This information allows the mapping function to have better coverage over the CLIP latent space.
To extract this information, we utilize two multimodal loss functions: (1) \textbf{cycle-consistency} loss; and (2) a novel \textbf{prompt-guided knowledge distillation} loss. We now cover our loss functions in further detail.
\vspace{-3mm}
\subsubsection{Reconstruction Loss}
\vspace{-1mm}
The simple and surprisingly effective approach taken in \cite{t2c2023} is to minimise the reconstruction loss of the mapped vision encoder and the CLIP vision encoder. The reconstruction loss is calculated using Mean Squared Error (MSE) as,  
\begin{equation}
    \label{eq:mse}
    \mathcal{L}_{mse} = \mathbb{E}_{\Mat{x} \sim \mathcal{D}^{vision}_{train}} \left[\norm{h(V_s(\Mat{x})) - V_t(\Mat{x})}_2^2\right]\ ,
\end{equation}
where $\mathcal{D}_{train}$ represents the training dataset. This guides $h$ to map the source vision subspace into the CLIP vision subspace.
\vspace{-2mm}


\vspace{-1mm}
\subsubsection{Aligning Variance in the Latent Spaces}
\vspace{-1mm}
Another important step shown in \cite{t2c2023} was re-scaling the variance of the CLIP and source vision encoders latent subspace, such that they are the same. From our observation, the variance alignment and re-scaling are crucial to ensure quick convergence.
We perform variance re-scaling of $V(\Mat{x})$ to $\hat{V}(\Mat{x})$ using,
\begin{equation}
    \label{eq:var1}
    var(V, \mathcal{D}) = \frac{\sum_{x \in \mathcal{D}} V( \Mat{x})^2}{\abs{\mathcal{D}}} - \left( \frac{\sum_{x \in \mathcal{D}} V( \Mat{x})}{\abs{\mathcal{D}}} \right) ^2 ,
\end{equation}
\vspace{-2mm}
\begin{equation}
\label{eq:var2}
    \hat{V}(\Mat{x}) = \sqrt{\frac{var(V, \mathcal{D})}{var_{target}}} \times V( \Mat{x}), 
\end{equation}
where $\mathcal{D}$ is the image dataset, $\Mat{x}$ is a single image such that $\Mat{x} \in \mathcal{D}$ and $var_{target}$ is the desired variance which both latent spaces are re-scaled to match.

As stated in \cite{t2c2023}, rescaling the variance of the features is important because some vision encoders embed inputs into low-variance spaces, which degrades the performance of the mapping functions due to the precision in the computations.
\vspace{-3mm}
\subsubsection{Multimodal Loss}
\vspace{-1mm}
The goal of the multimodal loss is to ensure that the mapping function learns information from the text subspace and the relationships between these two modalities.
However, the text features only exist within the CLIP latent space.
This is unlike the vision modality where we can map between the source vision encoder, \sourcevisenc, and the CLIP vision encoder, \targetvisenc. 
To address this, we define the inverse mapping function, $h^{-1}$, for mapping from the CLIP $d$-dimensional latent space into the $m$-dimensional source latent space, $h^{-1}: \mathbb{R}^d \mapsto \mathbb{R}^m$. Using the inverse mapping we enable $h$ to learn the text subspace and the relationships between text and vision.

We extract this information with: (1) Cycle-consistency loss; and (2) Prompt-guided knowledge distillation (PG-KD) loss. The cycle-consistency loss aims to ensure a data point from one domain is still consistent after it is mapped into the other domain and mapped back into its original domain. Additionally, it allows us to map text features between the two latent spaces. PG-KD utilizes the text prompts as zero-shot classifiers to extract the relationships between the text and image features. Whilst the cycle-consistency loss has been widely used in the community~\cite{wang2019cycletracking, CycleGAN2017, zhang2021cyclesegmentation, liu2020instancesegmentation, jeon2021temporalcorrespondence, shen2022siamese}, to our knowledge we are the first to propose PG-KD.

\noindent
\textbf{Cycle-consistency loss - } We apply cycle-consistency across the two latent spaces, in three subspaces: (1) the source vision encoder subspace, \sourcevisenc, (2) the CLIP vision encoder subspace, \targetvisenc, and (3) the CLIP text encoder subspace, \targettextenc. Specifically, features extracted by any encoder should be consistent with their original form, when mapped to and from the opposing latent space (\eg $V_s(\Vec{x}) = h^{-1} ( h ( V_s(\Vec{x}) )$).
Thus, our cycle-consistency loss is formulated as follows,

\vspace{-5mm}
\begin{equation}
    \label{eq:lcyc}
    \begin{split}
        \mathcal{L}_{cyc}  & = \mathbb{E}_{ \Vec{x} \sim \mathcal{D}^{vision}_{train}} [\| h ^{-1} (h (V_s (\Vec{x}))) - V_s ( \Vec{x} ) \|_1] \\
                                            & + \mathbb{E}_{ \Vec{x} \sim \mathcal{D}^{vision}_{train}} [\| h ( h^{-1} (V_t (\Vec{x} ) ) - V_t ( \Vec{x}) \|_1 ]  \\
                                            & + \mathbb{E}_{ \Vec{p} \sim \mathcal{D}^{text}_{train}} [\| h ( h ^ {-1} (T_t (\Vec{p} ) ) - T_t ( \Vec{p}) \|_1 ]  .
    \end{split}
\end{equation}


\noindent
\textbf{Prompt-Guided Knowledge Distillation - } Another avenue to extract the relationship between text and vision subspaces is by looking at how CLIP functions as a VLM. Specifically, CLIP classifies  $\Vec{x}$ as belonging to $\Vec{p}$ through the cosine similarity of $V_t(\Vec{x})$ and $T_t(\Vec{p})$ as, 
\vspace{-3mm}
\begin{equation}
\label{eq:cos}
    cos(V(\Mat{x}), T(\Vec{p})) = \frac{V(\Mat{x}) \cdot T(\Vec{p})}{\norm{V(\Mat{x})} \norm{T(\Mat{p})}}, 
\end{equation} 

\begin{equation}
\label{eq:softmax}
    \sigma(\Mat{z}) = \frac{e^{z_i}}{\sum_{j=1}^C e^{z_j}},
\end{equation}

\begin{equation}
\label{eq:zero-shot}
    S(V(\Mat{x}), T(\Vec{p})) = \sigma(cos(V(\Mat{x}), T(\Vec{p}))),
\end{equation}

\noindent
where $S$ is a zero-shot classifier. We consider four variants of the zero-shot classifier: (1) $S_t(V_t ( \Mat{x}), T_t( \Vec{p} ) )$; (2) $S_1 ( h ( V_s (\Mat{x}) ) , T_t ( \Vec{p} ) )$; (3) $S_2 ( V_s (\Mat{x}) , h^{-1}( T_t ( \Vec{p} ) ) )$; and (4) $S_3 ( h^{-1} ( V_t ( \Mat{x}) ),  h^{-1} ( T_t( \Vec{p} ) ) )$.
Note that variant 1, $S_t$ is the vanilla CLIP zero-shot classifier operating in the CLIP latent space.
Ideally, we want the zero-shot classifiers $S_1$, $S_2$, and $S_3$ to have the same output as  $S_t$. As such we address this by adapting Knowledge Distillation (KD)~\cite{distill2014}. That is, $S_t$ is considered as the teacher model and $S_1$, $S_2$, and $S_3$ are considered as the student models. 
We refer to this as the Prompt-Guided Knowledge Distillation (PG-KD). Standard KD~\cite{distill2014} utilises cross entropy with high temperature ($\approx20$) as it produces a softer probability distribution over the classes. They show that logit matching ($\ell_1$ distance) is a special case of knowledge distillation. We consider both cross-entropy with high temperature and the logit matching (\ie $\ell_1$ distance) methods.
The PG-KD loss function is defined as, 
\vspace{-2mm}
\begin{equation}
\label{eq:pg-kdLoss}
    \mathcal{L}_{pg-kd} = \sum_{i=1}^3 \mathbb{E}_{\mathcal{D}^{vision}_{train}, \mathcal{D}^{text}_{train}} [ d (S_t , S_i   ) ] , 
\end{equation}
\noindent
where $d$ is either the high-temperature cross-entropy or $\ell_1$ distance. We later investigate the optimal metric of $d$ in section \ref{sec:loss} of our ablations, finding $\ell_1$ to be optimal.

\vspace{-1mm}
\section{Experiments}
\vspace{-1mm}
\label{sec:results}
We first describe our training setup, selected datasets, selected vision encoders, zero-shot test setting and our baseline methods before providing a discussion and analysis of our results.

\begin{table*}
\resizebox{\textwidth}{!}{%
\begin{tabular}{@{}llllllllllll@{}} \toprule
          & \multicolumn{2}{l}{MobileNetV3~\cite{Howard2019SearchingFM}} & \multicolumn{2}{l}{DenseNet-121~\cite{Huang2016DenselyCC}} & \multicolumn{2}{l}{ResNet-18~\cite{He2015DeepRL}} & \multicolumn{2}{l}{DINOv1~\cite{caron2021emerging}} & \multicolumn{2}{l}{DINOv2~\cite{dinov22023}}  & CLIP~\cite{clip2021}\\ \cmidrule(lr){2-3} \cmidrule(lr){4-5} \cmidrule(lr){6-7} \cmidrule(lr){8-9} \cmidrule(lr){10-11} \cmidrule(lr){12-12}
          
Test Dataset   & LA~\cite{t2c2023} & Ours & LA & Ours & LA & Ours & LA & Ours & LA & Ours \\ \midrule
CIFAR-10~\cite{krizhevsky_learning_2009}& 50.93  & \textbf{63.7} \textcolor{Green}{+12.77}  & 61.46 & \textbf{64.51} \textcolor{Green}{+3.05} & 55.12 & \textbf{62.63} \textcolor{Green}{+7.51} & 72.97 & \textbf{77.52} \textcolor{Green}{+4.55} & 93.8 & \underline{\textbf{94.35}} \textcolor{Green}{+0.55} & 90.56  \\
CIFAR-100~\cite{krizhevsky_learning_2009}& 21.05 & \textbf{27.19} \textcolor{Green}{+6.14} & 28.25 & \textbf{30.75} \textcolor{Green}{+9.67} & 23.38 & \textbf{26.77} \textcolor{Green}{+3.39}& 41.18 & \textbf{42.68} \textcolor{Green}{+1.5}& 64.47 & \textbf{65.42} \textcolor{Green}{+0.95}& \underline{65.92}   \\
ImageNet-100~\cite{deng_imagenet_2009}& 48.28 & \textbf{57.95} \textcolor{Green}{+9.67}& 70.14 & \textbf{72.7} \textcolor{Green}{+2.56}& 63.36 & \textbf{70.78} \textcolor{Green}{+7.42}& 77.86 & \textbf{78.62} \textcolor{Green}{+0.76}& 85.34 & \textbf{86.96} \textcolor{Green}{+1.62}& \underline{87.48}  \\
ImageNet-1000~\cite{deng_imagenet_2009}&18.79&\textbf{26.78} \textcolor{Green}{+7.99}&37.43&\textbf{42.24} \textcolor{Green}{+4.81}&31.28&\textbf{40.58} \textcolor{Green}{+9.3}&49.02&\textbf{54.26} \textcolor{Green}{+5.24}&59.67&\textbf{63.62} \textcolor{Green}{+3.65}&\underline{65.39}\\
CUB-200~\cite{WahCUB_200_2011}& 2.95 & \textbf{12.78} \textcolor{Green}{+9.83}& 5.5 & \textbf{7.21} \textcolor{Green}{+1.71}& 5.4 & \textbf{14.68} \textcolor{Green}{+9.28}& 11.94 & \textbf{19.95} \textcolor{Green}{+8.01}& 18.74 & \textbf{29.32} \textcolor{Green}{+10.58}& \underline{54.28}   \\
Flowers-102~\cite{Nilsback08}& 3.26 & \textbf{7.38} \textcolor{Green}{+4.12}& 4.14 & \textbf{4.35} \textcolor{Green}{+0.21}& 2.74 & \textbf{6.68} \textcolor{Green}{+3.94}& 6.42 & \textbf{13.69} \textcolor{Green}{+7.27}& 16.55 & \textbf{25.32} \textcolor{Green}{+8.77}& \underline{71.24}  \\
Herbarium-19~\cite{tan2019herbarium} & \underline{\textbf{0.26}} & 0.22 \textcolor{BrickRed}{-0.04}& 0.11 & 0.11 & \textbf{0.11} & 0.03 \textcolor{BrickRed}{-0.08}& 0.187 & 0.187 & 0.037 & \textbf{0.187} \textcolor{Green}{+0.15}& 0.037 \\
Oxford-IIIT Pets~\cite{parkhi12a} & 46.69 & \textbf{64.24} \textcolor{Green}{+17.55}& 61.57 & \textbf{69.55} \textcolor{Green}{+7.98}& 62.36 & \textbf{72.6} \textcolor{Green}{+10.24}& 74.62 & \textbf{77.84} \textcolor{Green}{+10.24}& 80.05 & \textbf{86.96} \textcolor{Green}{+6.91}& \underline{89.71} \\
Stanford Cars~\cite{6755945} & 1.22 & \textbf{1.96} \textcolor{Green}{+0.74}& 1.1 & 1.1 & \textbf{1.72} & 1.59 \textcolor{BrickRed}{-0.62}& 1.47 & 1.47 & 3.31 & \textbf{5.28} \textcolor{Green}{+1.97}& \underline{63.51}  \\ \midrule
Average &21.49 & \textbf{29.13} \textcolor{Green}{+7.64} & 29.96 & \textbf{32.02} \textcolor{Green}{+2.06} & 27.27 & \textbf{32.93} \textcolor{Green}{+5.66} & 37.3 & \textbf{40.69} \textcolor{Green}{+3.39} & 43.21 & \textbf{50.86} \textcolor{Green}{+7.65} & \underline{65.35}\\
\bottomrule
\end{tabular}%
}
\vspace{-2mm}
\caption{Top-1 zero-shot test accuracy using CLIP text features and mapping the vision encoder features into the CLIP latent space. We compare our results against the recent SOTA \cite{t2c2023} which we refer to as LA for Linear Aligner. Our method achieves improved performance on nearly all datasets and model combinations. With DINOv2, we are even able to surpass the original performance of CLIP~\cite{clip2021} on CIFAR10 and 100 \cite{krizhevsky_learning_2009}. DINOv1 is a ViT-B/16 encoder. DINOv2 is a ViT-B/14 encoder. For CLIP, we use the ViT-B/16 vision encoder. The best results for each comparison are \textbf{bolded}, and the best overall zero-shot accuracy is \underline{underlined}. The delta in performance, shown as the (\textcolor{Green}{+}/\textcolor{BrickRed}{-}) beside each value, is with respect to the Linear Aligner top-1 zero-shot test accuracy for the same vision encoder.}
\vspace{-3mm}
\label{tab:zeroshot}
\end{table*}

\subsection{Setup Details}
\vspace{-1mm}
\noindent
\textbf{Training Setup - } We train our linear mapping functions $h$, and $h^{-1}$ on 20\% of the ImageNet~\cite{deng_imagenet_2009} training split without class labels/groundtruth. 
We use only 20\% of the training split instead of 100\% for two reasons: firstly to limit the computational requirement of training, and secondly, we find it delivers optimal performance, as reported in Table Table~\ref{tab:limited_data_ours}. Additionally, Moayeri~\etal \cite{t2c2023} substantiate this result in their supplementary material A.1.
We train for only a single epoch using the Adam~\cite{adamoptim} optimizer with an initial learning rate of 0.0001 and utilize a cosine annealing learning rate decay~\cite{cosineLR17}. When training, our overall loss function consits of the summation of our individual loss functions discussed in Section \ref{sec:method}.  In accordance with \cite{t2c2023} we re-scale both latent spaces to a target variance of 4.5. 
Unless otherwise stated, the zoom-shot results in this section use the logit matching variant for the PG-KD loss. The comparisons between the logit matching and the High Temperature Cross Entropyvariants are discussed in the ablation results in Section~\ref{sec:ablations}.
 
As the Zoom-shot training does not use any class labels/groundtruth or text-image paired data, we generated a set of general prompts using ChatGPT~\cite{ChatGPT}. Examples of these prompts include ``a photo of a dog", ``a photo of a building", and ``a photo of a computer". We intuitively decided to keep these prompts general in the hope to capture different regions of the CLIP text subspace. In total, there are 50 general prompts used. The details of the prompts are shown in the supplementary material.

\noindent
\textbf{Datasets - }We test our Zoom-shot approach across eight classification datasets of varying granularity. Specifically, we use ImageNet~\cite{deng_imagenet_2009}, CIFAR-10 and CIFAR-100~\cite{krizhevsky_learning_2009} as the `coarse' or general classification datasets as their classes are more broadly defined and should be easier to classify. In addition to these, we use five `fine-grained' datasets of varying difficulty, these being CUB-200~\cite{WahCUB_200_2011}, Flowers-102~\cite{Nilsback08}, Herbarium-19~\cite{tan2019herbarium}, Oxford-IIIT pets~\cite{parkhi12a} and Stanford-Cars~\cite{6755945}. The Herbarium-19 dataset is by far the most challenging with 683 classes, where each class is an individual species of the same flowering plant family. We use the pre-defined test splits for each dataset when reporting the top-1 test accuracy throughout this work.

\noindent
\textbf{Vision Encoders - } We use Zoom-shot to transfer the zero-shot capabilities of CLIP to five different vision encoders of varying size and complexity. These are MobileNetV3 small\cite{Howard2019SearchingFM}, DenseNet121~\cite{Huang2016DenselyCC}, ResNet18~\cite{He2015DeepRL}, DINOv1 ViT-B/16~\cite{caron2021emerging}, DINOv2 ViT-B/14~\cite{dinov22023}. All vision encoders are initialized using pre-trained ImageNet weights. We selected this range of models as they offer a meaningful trade-off of strengths and weaknesses, demonstrating the expected performance of our approach across a variety of potential use cases ranging from cloud-based to edge-based applications.

\noindent
\textbf{Zero-shot Setting - } We measure the top-1 zero-shot classification accuracy of our vision encoders using a set of prompts. These prompts are constructed by including the class name $C$ in a prompt template, such as ``An image of a \{$C$\}". We follow the zero-shot setting used in CLIP~\cite{clip2021}, where multiple prompt templates are averaged for an individual class. Additionally, we use the prompt templates from \cite{clip2021} where available for our datasets. More details on the exact prompts used can be found in the supplementary material.

\noindent
\textbf{Baseline methods - } We compare our results against two baseline methods. The first is the recent state-of-the-art method proposed by Moayeri \emph{et al.}~\cite{t2c2023}, which we refer to as \textbf{Linear Aligner (LA)} in all results. We use their training code as provided on their GitHub\footnote{\url{https://github.com/k1rezaei/Text-to-concept/}} with their described optimal training settings. We also use CLIP~\cite{clip2021} as an additional comparison. CLIP should serve as the upper bound of performance for Zoom-shot. We use the ViT-B/16 CLIP image encoder for all results. Both baseline methods follow the same zero-shot settings as described above.

\subsection{Zero-shot Classification}
\vspace{-1mm}
Here we discuss our main results under two core settings: (1) mapping features from the source vision encoder across to the CLIP latent space, and (2) mapping the text features from the CLIP text encoder across to the $m$-dimensional source latent space. Overall, Zoom-shot, with its multimodal loss in addition to the reconstruction loss, provides a significant improvement over the Linear Aligner method.


\noindent
\textbf{Mapping Vision Encoder Features - } Table \ref{tab:zeroshot} shows our main results comparing the top-1 zero-shot accuracy of our selected models and datasets. These results serve as a direct comparison to the Linear Aligner (LA) method~\cite{t2c2023}. We see a consistent improvement on nearly every model and dataset tested. The most significant improvements can be seen with MobileNetV3 small on the coarse datasets, and DINOv2 with the fine-grained datasets. These results show that the proposed multimodal losses, which exploit the relationship between text and vision, are crucial for enabling the linear mapping function, $h$, to more accurately capture CLIP's latent space. This in effect provides more accuracy for the smaller models with a less discriminative feature space; while at the same time being able to take advantage of the more discriminative latent spaces such as DINO v1 and v2. 
Surprisingly, both LA and our method beat CLIP's zero-shot accuracy on CIFAR-10. We conjecture this occurs due to the quality of DINOv2 as a vision encoder compared to the CLIP ViT-B/16 vision encoder. 
Once the mapping function is trained, it is mapping more accurate image features into CLIP latent space. These mapped features are more consistent with the text prompts in comparison to the features produced by CLIP's vision encoder. Our method also comes within a percentage point on the other coarse datasets. 

We observe that every model underperforms on the fine-grained datasets, excluding Oxford Pets and Herbarium-19, with respect to CLIP. The most significant performance drop can be seen on Stanford Cars. 
One possible explanation is that this may be caused by the training not adequately covering the regions of the latent space that relate to this dataset. For the Herbarium dataset, we see poor performance across all models, including CLIP. In fact, most of the targeted vision encoders actually outperform CLIP. Overall, we attribute the poor performance on this dataset as a limitation of the zero-shot prompting method. The Herbarium dataset contains very fine-grained classes, and the class names alone do not provide adequate detail in order to distinguish between the images for each class.

\noindent
\textbf{Mapping CLIP Text Features - } Our method allows us to perform zero-shot classification in the $m$-dimensional source latent space by mapping the CLIP text features using the inverse mapping function, $h^{-1}$. This could be beneficial for low-powered limited-computing applications as no extra mapping is required for each image to perform the zero-shot classification. Table~\ref{tab:vis_enc_zeroshot} contains our results for mapping text features from the CLIP latent space, into the $m$-dimensional source latent space. We see a reduction in the top-1 zero-shot accuracy for every model, although some models and datasets still achieve competitive performance in comparison to the LA method. We posit that the reduction in performance is due to the previously discussed modality gap~\cite{liang2022mind, shi2023towards}. The existence of the modality gap suggests that the mapping function needs to specifically learn information from each subspace to perform optimally. 
Our proposed zoom-shot method does not specifically aim to optimize the text subspace within the $m$-dimensional source latent space.

\begin{table}[]
\resizebox{\linewidth}{!}{%
\begin{tabular}{@{}llllll@{}} \toprule
Test Dataset         & MobileNetV3 & DenseNet-121 & ResNet-18 & DINOv1 & DINOv2 \\ \midrule
CIFAR-10        & 55.55 \textcolor{Green}{+4.62}& 58.57 \textcolor{BrickRed}{-2.89} & 55.01 \textcolor{BrickRed}{-0.11}& 79.31 \textcolor{Green}{+6.34}& 85.98 \textcolor{BrickRed}{-7.82}\\
CIFAR-100       & 21.34 \textcolor{Green}{+0.29}& 24.13 \textcolor{BrickRed}{-4.12}& 18.06 \textcolor{BrickRed}{-5.32}& 41.91 \textcolor{Green}{+0.73}& 63.29 \textcolor{BrickRed}{-1.18}\\
ImageNet-100    & 30.28 \textcolor{BrickRed}{-18.0}& 60.32 \textcolor{BrickRed}{-9.82}& 51.86 \textcolor{BrickRed}{-11.5}& 73.98 \textcolor{BrickRed}{-3.88}& 82.0 \textcolor{BrickRed}{-3.34}\\
ImageNet-1000   & 11.7 \textcolor{BrickRed}{-7.09}&35.5 \textcolor{BrickRed}{-1.93}&28.13 \textcolor{BrickRed}{-3.15}& 47.63 \textcolor{BrickRed}{-1.39}& 60.78 \textcolor{Green}{+1.11}\\ 
CUB-200         & 1.48 \textcolor{BrickRed}{-1.47}& 5.47 \textcolor{BrickRed}{-0.03}& 3.38 \textcolor{BrickRed}{-2.02}& 17.65 \textcolor{Green}{+5.71}& 32.41 \textcolor{Green}{+13.67}\\
Flowers-102     & 5.05 \textcolor{Green}{+1.79}& 6.73 \textcolor{Green}{+2.38}& 6.48 \textcolor{Green}{+3.74}& 11.04 \textcolor{BrickRed}{-2.65}& 25.35 \textcolor{BrickRed}{+8.8}\\
Herbarium-19    & 0.41 \textcolor{Green}{+0.15}& 0.149 \textcolor{Green}{+0.039}& 0.187 \textcolor{Green}{+0.077}& 0.112 \textcolor{BrickRed}{-0.075}& 0.112 \textcolor{Green}{+0.075}\\
Oxford-IIIT Pets& 14.93 \textcolor{BrickRed}{-32.03}& 52.11 \textcolor{BrickRed}{-9.46}& 47.91 \textcolor{BrickRed}{-14.45}& 76.64 \textcolor{Green}{+2.82}& 82.82 \textcolor{Green}{+2.77}\\
Stanford Cars   & 0.85 \textcolor{BrickRed}{-0.37}& 1.35 \textcolor{Green}{+0.25}& 1.35 \textcolor{BrickRed}{-0.37}& 1.72 \textcolor{Green}{+0.25}& 4.3 \textcolor{Green}{-0.99}\\   \midrule
Average & 15.73 \textcolor{BrickRed}{-5.79}& 29.15 \textcolor{BrickRed}{-0.81}& 23.6 \textcolor{BrickRed}{-3.36}& 38.89 \textcolor{Green}{+1.59}& 48.56 \textcolor{Green}{+5.35}
\\
\bottomrule
\end{tabular}%
}
\caption{Top-1 zero-shot test accuracy when using CLIP~\cite{clip2021} text features mapped into the $m$-dimensional source latent space. Despite weaker performance, some variants still have competitive performance. Notable improvements can be observed in the Herbarium-19 dataset~\cite{tan2019herbarium}. The delta in performance, shown as the (\textcolor{Green}{+}/\textcolor{BrickRed}{-}) beside each value, is with respect to the Linear Aligner~\cite{t2c2023} top-1 zero-shot test accuracy from Table \ref{tab:zeroshot}.}
\vspace{-3mm}
\label{tab:vis_enc_zeroshot}
\end{table}

\vspace{-2.5mm} 
\section{Ablations}
\label{sec:ablations}
\vspace{-1mm}
In this section, we investigate a number of ablation experiments in order to fully understand and explain: (1) the impact of the different loss functions and how they benefit the mapping task, (2) how the size of the training dataset and amount of training steps performed impacts performance, and (3) how the distribution of training images impacts performance. For all ablation studies, we use the MobileNetV3 small model. The results for the other vision encoder models are available in the supplementary material. 

\subsection{Impact of Loss Functions}
\label{sec:loss}
Firstly, we investigate the performance of using high temperature cross entropy (HT-CE) and logit matching (LM) to learn the mapping functions $h$ and $h^{-1}$. HT-CE achieves 17.04\% zero-shot accuracy on CIFAR-10~\cite{krizhevsky_learning_2009} and 1.79\% on CIFAR-100~\cite{krizhevsky_learning_2009} for mapping features with $h$. LM nearly doubles this performance with 30.42\% on CIFAR-10 and 4.06\% on CIFAR-100. Additionally, using mapping function $h^{-1}$ HT-CE achieves 9.75\% and 1.72\% on CIFAR-10 and 100, respectively, while LM achieves 12.59\% and 1.81\%. We provide additional experiments regarding this point in the supplementary material. This leads us to select logit matching as our preferred form of knowledge distillation in PG-KD. 

Next, we investigate the impact of the loss functions used, namely, reconstruction loss as Mean Squared Error (MSE), PG-KD with logit matching and Cycle Consistency (CC) loss functions have on the performance of the mapping function. In Table~\ref{tab:loss_performance} we test the performance of each loss function in isolation and in different combinations.
We observe that MSE alone has a significant contribution to the overall performance, while CC performs the worst in isolation.
When they are paired, MSE+PG-KD result in a reduction in performance whereas MSE+CC result in a slight improvement. Additionally, the mapping function struggles to learn any meaningful mapping between the latent spaces when it is trained by using CC+PG-KD. This suggests that MSE is a crucial loss function which corroborates the finding in~\cite{t2c2023}. However, combining MSE with both CC and PG-KD significantly boosts the performance over the MSE-only variant. This highlights the importance of the mapping function to learn the information from both text and vision subspace and the relationship between them.


\begin{table}[]
\resizebox{\linewidth}{!}{%
\begin{tabular}{@{}llllllll@{}} \toprule
Dataset   & MSE & MSE+PG-KD & MSE+CC & PG-KD & CC & CC+PG-KD & All \\ \midrule
CIFAR-10   & 58.02 & 52.58 & 59.3 & 36.62 & 12.48  & 13.97 & 63.7 \\
CIFAR-100  & 23.36 & 23.32 & 27.1 & 8.3 & 0.68 & 0.86 & 27.19\\
ImageNet-100  & 54.91  & 48.04 & 57.06& 4.24 & 0.66  & 0.86 & 57.95 \\
ImageNet-1000  & 20.51 & 20.61 & 26.52& 0.122 &  0.104  &  0.118 & 26.78  \\
CUB-200    & 3.98  & 4.00  & 12.2& 0.6 & 0.27 & 0.24 & 12.78 \\
Flowers-102   & 3.1  & 3.1  & 7.6& 0.53 & 1.0 & 1.33 & 7.38\\
Herbarium-19 & 0.37 & 0.37  & 0.18& 0.07 & 0.11 & 0.149 & 0.22 \\
Oxford-IIIT Pets    & 49.79 & 49.74 & 63.54 & 5.75 & 1.98  & 1.96 & 64.24  \\
Stanford Cars  & 1.22  & 1.22  & 2.21 & 0.73 & 0.73  & 0.61 & 1.96\\ \midrule
Average & 23.92 & 22.54 & 28.42 & 6.33 & 2.01 & 2.23 & 29.13 \\
\bottomrule
\end{tabular}%
}
\vspace{-2mm}
\caption{Top-1 test accuracy of each individual loss function on MobileNetV3 small~\cite{Howard2019SearchingFM}. MSE: Mean Squared Error; PG-KD: Prompt-Guided Knowledge Distillation with the logit matching variant; CC: Cycle-consistency. The results show that combining all the loss functions yields significant improvement over the MSE loss alone which is used in the recent state-of-the-art Linear Aligner method ~\cite{t2c2023}.}
\vspace{-2mm}
\label{tab:loss_performance}
\end{table}


\begin{table}[]
\centering
\resizebox{0.8\linewidth}{!}{%
\begin{tabular}{@{}llllll@{}} \toprule
Dataset & 100\% & 20\% & 5\% & 1\% & 1\% (20 epochs) \\ \midrule
CIFAR10 & 59.44 & 63.7 & 58.27 & 40.95 & 61.47 \\
CIFAR100 & 26.59 & 27.19 & 24.84 & 11.14 & 26.65 \\
ImageNet-100 & 61.1 & 57.95 & 50.94 & 28.64 & 55.44 \\
ImageNet-1000 & 27.29 & 26.78 & 21.01 & 6.23 & 24.53 \\
CUB200 & 12.47 & 12.78 & 8.54 & 1.88 & 11.32 \\
Flowers & 6.89 & 7.38 & 4.56 & 3.69 & 7.32 \\
Herbarium & 0.41 & 0.22 & 0.223 & 0.223 & 0.41 \\
Oxford & 64.94 & 64.24 & 54.84 & 17.14 & 61.78 \\
Stanford & 1.96 & 1.96 & 0.85 & 0.49 & 1.22 \\ \midrule
Average & 29.01 & 29.13 & 24.90 & 12.26 & 27.78\\
\bottomrule
\end{tabular}%
}
\vspace{-2mm}
\caption{Top-1 test accuracy using different amounts of ImageNet~\cite{deng_imagenet_2009} training data. In the final column, we show that simply extending the training time to 20 epochs allows our method to nearly match the performance at 20\% training data. All other columns are trained for 1 epoch.
}
\vspace{-2mm}
\label{tab:limited_data_ours}
\end{table}

\begin{table}[]
\resizebox{\linewidth}{!}{%
\begin{tabular}{@{}lllllll@{}} \toprule
Dataset & \makecell{100\%\\(6 epochs)}&\makecell{20\%\\(6 epochs)}&\makecell{1\%\\(6 epochs)}&\makecell{1\%\\(20 epochs)}& \makecell{1\%\\(120 epochs)} \\ \midrule
CIFAR10 & 47.3& 50.93 &45.84 & 50.34 & 51.83 \\
CIFAR100 & 19.85& 21.05 &13.16 & 18.64 & 21.09 \\
ImageNet-100 & 51.96& 48.28 &35.78 & 47.56 & 52.74 \\
ImageNet-1000 & 17.7& 18.79 &13.61 & 15.34 & 18.54 \\
CUB200 & 2.39& 2.95 &1.43 & 2.27 & 3.12  \\
Flowers & 3.11& 3.26 &2.06 & 2.82 & 3.69  \\
Herbarium & 0.0373& 0.26 &0.335 & 0.373 & 0.335 \\
Oxford & 44.37& 46.69& 26.76 & 41.56 & 48.54  \\
Stanford & 1.23& 1.22 &1.11 & 0.614 & 0.737  \\ \midrule
Average & 20.88& 21.49 & 15.57 & 19.95 & 22.29 \\
\bottomrule
\end{tabular}%
}
\vspace{-2mm}
\caption{Top-1 test accuracy of Linear Aligner~\cite{t2c2023} using different amounts of ImageNet~\cite{deng_imagenet_2009} training data. In the final column, we show that simply extending the training time to 120 epochs allows the method to nearly match the performance at 20\% training data.}
\vspace{-3mm}
\label{tab:limited_data}
\end{table}




\begin{figure}
    \centering
    \includegraphics[trim={11mm 10mm 11mm 2.5mm},width=0.6\linewidth,clip]{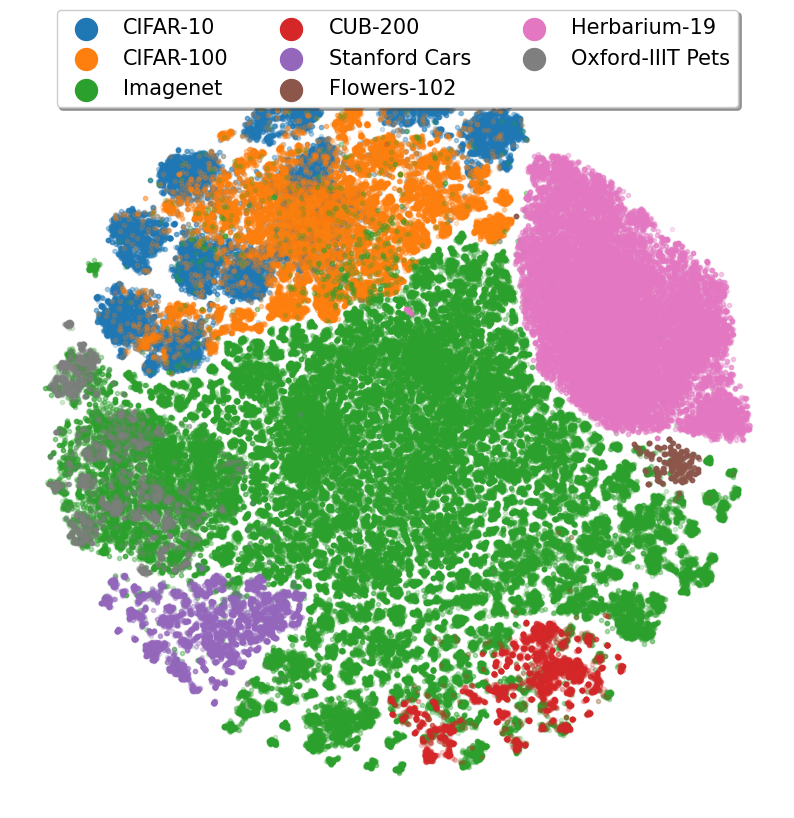}
    \vspace{-2mm}
    \caption{TSNE~\cite{TSNE} visualisation of all training datasets in the CLIP ViT-B/16 image encoders latent space. We can observe an interspersion of ImageNet features among the datasets like Oxford-IIIT pets and CUB-200; while at the same time there are differing degrees of separation for the Stanford Cars and Herbarium-19 datasets.}
    \label{fig:tsne_all_tests}
\end{figure}

\vspace{-1mm}
\subsection{Size of Training Dataset}
\vspace{-1mm}
\label{sec:ab_data}
In this set of experiments, we test the effects of lowering the size of the training dataset, presented in Table \ref{tab:limited_data_ours}. We find on average there is a slight drop in performance for training on the 100\% of the ImageNet data compared to training with only 20\% of the data.
As we drop the amount of training data below 20\% we see a subsequent drop in the zero-shot accuracy on each dataset. Lastly, we investigate whether this drop in accuracy is actually due to the reduction in data or to fewer training steps being performed at the lower data regimes. To test this, we increase the number of epochs at 1\% training data to 20, in order to match the effective number of training steps for a single epoch of training with 20\% training data. By doing this, we are able to regain a surprising amount of the zero-shot performance, which is competitive to the 20\% training data. This implies that in our context, the quantity of training steps holds greater significance than the volume of training data used. These results prompted us to investigate whether the recent state-of-the-art Linear Aligner method~\cite{t2c2023} exhibits similar properties. Interestingly, Table \ref{tab:limited_data} shows that 1\% of the training data with an equal amount of compute actually outperforms the results in Table~\ref{tab:zeroshot}. We compared 6 epochs because it was the optimal setting as stated in~\cite{t2c2023}; 20 epochs so we could directly compare against the results in Table~\ref{tab:limited_data_ours}; and 120 epochs so the compute scaling at 1\% training data matched the optimal 6 epochs at 20\%. 
Despite showing similar properties, the Linear Aligner method still performs significantly lower than the proposed Zoom-shot method.

\begin{table}[]

\resizebox{\linewidth}{!}{%
\begin{tabular}{@{}llllll@{}} \toprule
Testing Dataset & MobileNetV3 & DenseNet121 & ResNet18 & DINOv1 & DINOv2 \\
\midrule
CIFAR10 & 62.6 \textcolor{BrickRed}{-1.1} & 83.56 \textcolor{Green}{+19.05}& 90.56 \textcolor{Green}{+27.93}& 90.22 \textcolor{Green}{+12.7}& 95.39 \textcolor{Green}{+1.04}\\
CIFAR100 & 25.39 \textcolor{BrickRed}{-1.8} & 44.05 \textcolor{Green}{+13.3} & 42.7 \textcolor{Green}{+15.93}& 65.11 \textcolor{Green}{+22.43}& 75.35 \textcolor{Green}{+9.93}\\
CUB200 & 19.19 \textcolor{Green}{+6.41}& 34.33 \textcolor{Green}{+27.12}& 30.72 \textcolor{Green}{+16.04}& 46.65 \textcolor{Green}{+26.7}& 58.54 \textcolor{Green}{+29.22}\\
Flowers & 11.93 \textcolor{Green}{+4.55}& 43.75 \textcolor{Green}{+39.4}& 38.54 \textcolor{Green}{+31.86}& 58.43 \textcolor{Green}{+44.74}& 74.89 \textcolor{Green}{+49.57}\\
Herbarium & 0.111 \textcolor{BrickRed}{-0.109}& 0.0 \textcolor{BrickRed}{-0.11}& 0.0747 \textcolor{Green}{+0.045}& 0.0373 \textcolor{BrickRed}{-0.149}& 0.0747 \textcolor{BrickRed}{-0.112}\\
Oxford & 58.95 \textcolor{BrickRed}{-5.29}& 86.39 \textcolor{Green}{+16.84}& 83.24 \textcolor{Green}{+10.64}& 86.37 \textcolor{Green}{+8.53}& 90.37 \textcolor{Green}{+3.41}\\
Stanford & 4.91 \textcolor{Green}{+2.95}& 19.04 \textcolor{Green}{+17.94}& 14.37 \textcolor{Green}{+12.78}& 34.15 \textcolor{Green}{+32.68}& 58.47 \textcolor{Green}{+53.19}\\ \midrule 
Average & 26.14 \textcolor{Green}{+0.62}& 44.45 \textcolor{Green}{+19.08}& 42.89 \textcolor{Green}{+16.46}& 54.42 \textcolor{Green}{+21.09}& 64.73 \textcolor{Green}{+20.89}\\
\bottomrule
\end{tabular}%
}
\vspace{-2mm}
\caption{Top-1 accuracy from training the mapping functions on the training data of each testing dataset. We see massive improvements from training on an aligned distribution. These results show that the distribution of training images have a major impact on the transferred zero-shot performance. The delta in performance, shown as the (\textcolor{Green}{+}/\textcolor{BrickRed}{-}) beside each value, is with respect to our top-1 zero-shot test accuracy from Table \ref{tab:zeroshot}.}
\vspace{-5mm}
\label{tab:retrainWithOwnTraining}
\end{table}

\vspace{-1mm}
\subsection{Effect of the Training Distribution}
\vspace{-1mm}
In the previous section, all the results are produced by training the mapping function with images from ImageNet.
Despite significant improvement over the recent state-of-the-art, there is still a significant gap between the zero-shot performance of vision encoders augmented with Zoom-shot and the upper bound of CLIP. 
The significant gap is observable in the fine-grained classification datasets such as the Stanford car dataset.
This gap could be due to the specialized domain presented in these datasets which may not be well represented in the ImageNet dataset.
In this section, we study the effect of how the distribution of training images impacts this performance gap.
We show the TSNE visualization of the CLIP image encoder feature space across all the training sets in Figure \ref{fig:tsne_all_tests}. From this, we can see the images from ImageNet are most interspersed among the Oxford-IIIT Pets, CUB-200 and CIFAR datasets. 
This would explain why ImageNet training produces the best performance on these datasets, as the ImageNet images clearly cover the diverse area of the latent space, including regions of these aforementioned datasets.
However, the ImageNet images do not seem to cover the Stanford Cars and Herbarium-19 datasets to the same degree. We can see only a handful of Imagenet images interspersed among Stanford Cars, and no images amongst Herbarium-19.

To further investigate this, for each dataset we retrain the Zoom-shot method using the corresponding training images (again without labels).
Table~\ref{tab:retrainWithOwnTraining} shows that with the exception of the Herbarium-19 dataset, there is a significant increase in performance.
This suggests that the distribution of training images within the CLIP latent space has a significant impact on the quality of the mapping for images relating to those regions. Again, the poor performance on the Herbarium-19 might be caused by the poor CLIP's zero-shot performance in this dataset, as discussed in Section~\ref{sec:results}. Note that the results reported in Table~\ref{tab:retrainWithOwnTraining} suggest that the performance of the Zoom-shot method can be further boosted if it is trained with datasets that well represent the target domain. Further investigations on training the Zoom-shot method with large-scale datasets will be considered as future work.

\vspace{-2mm}
\section{Conclusion}
\vspace{-1mm}
In this work, we demonstrate the importance of capturing the interactions between text and image features, for cross-model alignment, when aligning a pre-trained vision encoder to CLIP's latent space. This is due to the recently identified modality gap, resulting in two distinct subspaces between the text and image features. 
As a result, learning from only a single modality falls short of capturing the entire latent space.
Our solution, Zoom-shot, addressed this with our multimodal loss functions: Cycle-Consistency, and Prompt-Guided Knowledge Distillation. Overall, this improves training efficiency, as Zoom-shot learns the linear mapping in only a single epoch. Furthermore, Zoom-shot utilises entirely unlabeled and unpaired data. Once learnt, the mapping augments the pre-trained vision encoders as zero-shot classifiers. In our ablation studies, we discovered Zoom-shot allows for a trade-off between data and compute. Additionally, we found the zero-shot performance varies depending on the distribution of training images.
We envisage the insights gleaned from our work will enable the adoption of large multimodal models for a range of novel applications, and facilitates the efficient development of these models for the broader research community.

\section*{Acknowledgement}
This work has been supported by the SmartSat CRC, whose activities are funded by the Australian Government’s CRC Program; and partly supported by Sentient Vision Systems. Sentient Vision Systems is one of the leading Australian developers of computer vision and artificial intelligence software solutions for defence and civilian applications.

{
    \small
    \bibliographystyle{ieeenat_fullname}
    \bibliography{main}
}

\clearpage
\setcounter{page}{1}
\maketitlesupplementary

In our supplementary material we present the following: (1) a visualization of the modality gap present in CLIP's latent space; (2) details pertaining to Zoom-shot's training loop, including PyTorch like pseudo code; (3) the Zoom-shot training prompts as used in Prompt-Guided Knowledge Distillation; (4) details relating to the zero-shot prompts we utilized for the zero-shot classification tests; and lastly, (5) additional ablation results across the vision encoders not presented in the main papers ablation studies. 

\section{Modality Gap Visualization}

\begin{figure}[h]
    \centering
    \includegraphics[trim={15mm, 15mm, 15mm, 15mm}, width=0.65\linewidth, clip]{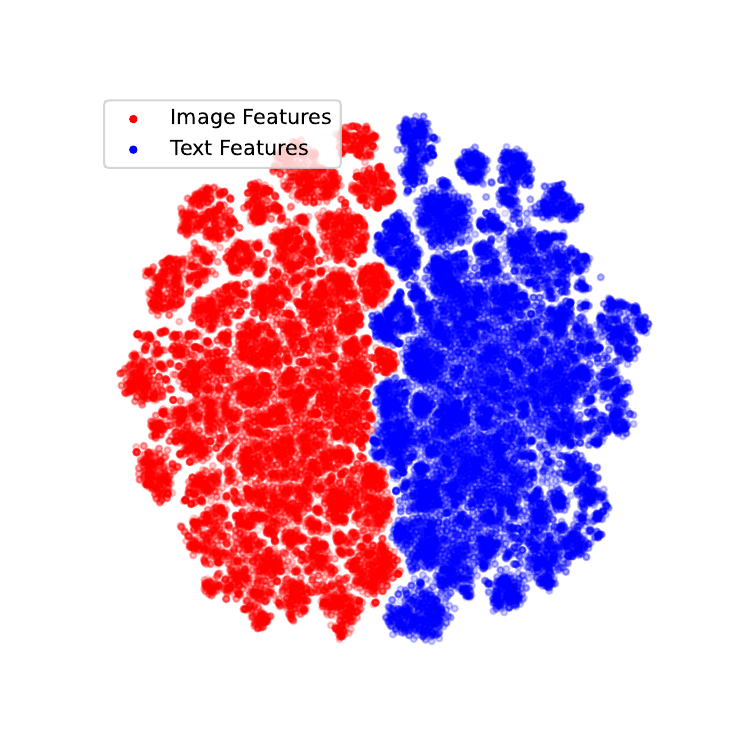}
    \caption{TSNE~\cite{TSNE} visualization of pretrained image and text features from a randomly selected 10\% of the MS-COCO dataset \cite{mscoco}. The divide between the sets of features demonstrates the modality gap clearly present in CLIP~\cite{clip2021}.}
    \label{fig:modality_gap}
\end{figure}

The modality gap is a recently discovered phenomena present in the latent space of CLIP~\cite{clip2021}. The gap occurs, as explained in \cite{liang2022mind, shi2023towards}, due to a local minima in the contrastive learning objective of CLIP's training. CLIP's training objective was devised to align text and image features from their respective encoder through the use of a join latent space, such that a paired text and image feature should share the same point in that latent space. In reality, the encoders map features to different subspaces of the latent space which share the semantic ordering of paired points. This is best shown in \cite{shi2023towards} (see the Figure 2 in their work), which demonstrates the local minima through the use of a hyperspherical latent space. We visualize the modality gap ourselves in Figure \ref{fig:modality_gap} through the use of TSNE dimensionality reduction~\cite{TSNE}. For this plot, we visualize image and text features using CLIP's ViT-B/16 image encoder and corresponding text encoder. For the data, we randomly selected 10\% of MS-COCO's~\cite{mscoco} image and text pairs. We use MS-COCO, instead of our tested datasets, as MS-COCO is an image captioning dataset, therefore; each caption should be more closely aligned to its corresponding image. Additionally, the captions provide more unique text features for the sake of this visualization. In Figure \ref{fig:modality_gap} we can observe a clear divide between the two modalities, demonstrating the modality gap.

\section{Training Loop}
We described our loss functions in Section \ref{sec:method} of the main paper, and attempted to articulate their uses in training. To reinforce this understanding we present Pytorch like pseudo code for Zoom-shot's training loop, shown in Algorithm \ref{alg:train_loop}, which we will now discuss. Firstly, after computing the initial mappings with $h$ and $h^{-1}$, we can use the reconstruction loss (MSE) between the outputs and the ground truths. We can also cycle each output through its opposite mapping function to obtain reconstructions of the original features. The cycle-consistency loss can then be used on these reconstructions. Additionally, we can cycle our CLIP text features through $h$ and $h^{-1}$ and compute the loss on the text features reconstruction. Lastly, with the different outputs and reconstructions of the features obtained, we can compute probability distributions for each pairing. Using Prompt-Guided Knowledge Distillation we can then calculate the loss between the computed distributions and the distribution of the original CLIP image and text features. After backpropagating the accumulated losses, this concludes a single training loop of our method.

\begin{algorithm}
[t]\captionsetup{labelfont={sc,bf}, labelsep=newline}
  \caption{Pytorch like pseudocode for a single training step when training the mapping functions. This demonstrates the numerous places where our loss functions can be utilized during training.}
  \label{alg:train_loop}

\begin{algorithmic} \\
\hspace*{-3mm}\includegraphics[trim={25mm 131mm 88mm 25mm}, width=\linewidth, clip]{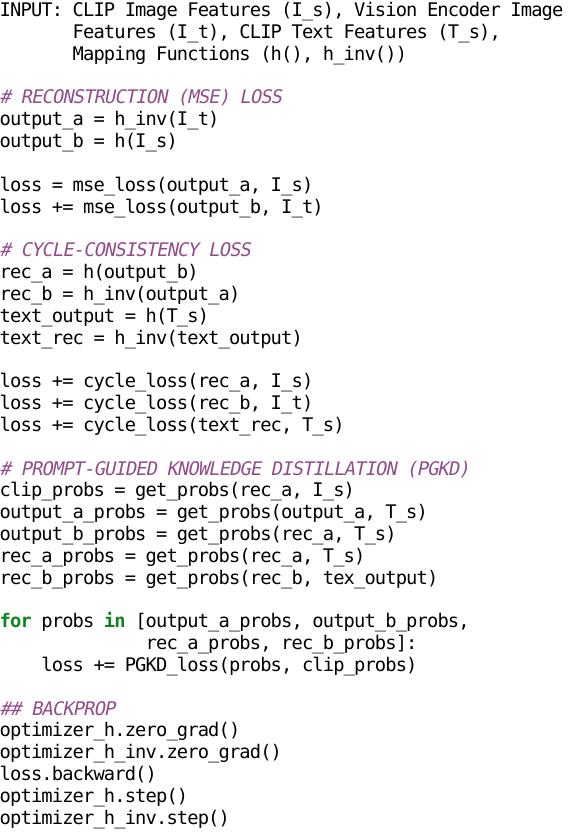}
\end{algorithmic}
\end{algorithm}

\vspace{-2mm}
\section{Training Prompts}
\begin{figure}[h]
    \centering
    \includegraphics[trim={25mm 180mm 90mm 25mm}, width=\linewidth, clip]{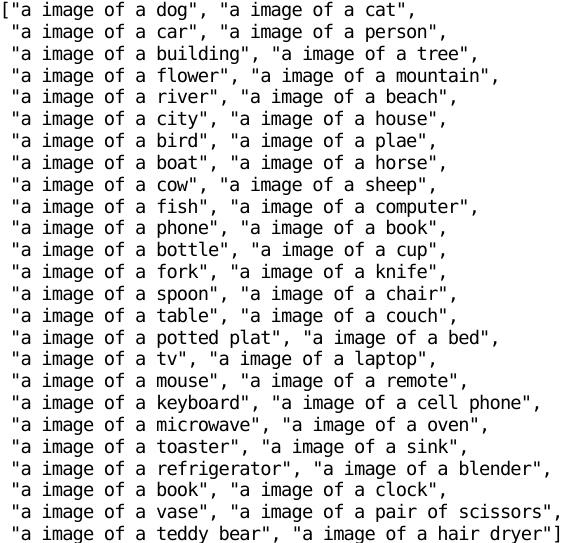}
    \caption{Prompts used for Prompt Guided Knowledge Distillation as generated by ChatGPT~\cite{ChatGPT}.}
    \label{fig:prompts}
\end{figure}

When training our Zoom-shot method, we utilize 50 general prompts for the Prompt-Guided Knowledge Distillation. As mentioned in Section \ref{sec:results}, these prompts were randomly generated using Chat-GPT~\cite{ChatGPT} with the following input ``Randomly generate 50 zero-shot prompts for me, in the format `an image of a \{\}', replace \{\} with a single word object or thing". This resulted in the list of prompts shown in Figure \ref{fig:prompts}.

\vspace{-2mm}
\section{Zero-shot Classification Prompts}
When testing our various vision encoders, under the zero-shot setting, we follow the practice used in CLIP \cite{clip2021}. Meaning, that where available, we include the class name inside either a single, or multiple, prompt templates. When using multiple prompt templates, the final text feature is simply the average of the text features produced by the prompts. The templates used for each dataset can be found on our GitHub.

\section{Ablations}
This section provides further ablation studies across our range of selected vision encoders not shown in the main paper. To summarize, we (1) conduct an analysis of high-temperature cross-entropy against logit matching for use in Prompt-Guided Knowledge Distillation; (2) demonstrate the impact of the loss functions through testing different combinations of the loss functions; and (3) investigate the performance obtained with different percentages of training data. These additional ablations are provided for the DenseNet121~\cite{Huang2016DenselyCC}, ResNet18~\cite{He2015DeepRL}, DINOv1 ViT-B/16~\cite{caron2021emerging} and DINOv2 ViT-B/14~\cite{dinov22023} vision encoders as we provided results for MobileNetV3 small in the main paper (Section \ref{sec:ablations}).


\subsection{HT-CE vs LM}
The results in the Section \ref{sec:loss} of the main paper show logit matching (LM) (e.g. $\ell_1$ distance) outperforms high-temperature cross-entropy (HT-CE). This informed our selection of LM for use in Prompt-Guided Knowledge Distillation. Here we present additional results. Table \ref{tab:pgkd_loss1} and \ref{tab:pgkd_loss2} compares HT-CE and LM across our range of selected datasets and vision encoders. Overall, these results reiterate our conclusion that LM outperforms HT-CE. LM achieves a higher average accuracy in eight of the ten comparison.
However, these results are more nuanced than the near doubling of performance as stated. In fact, this near doubling of performance appears to primarily occur on the CIFAR datasets. 


\begin{table*}[]
\centering
\begin{tabular}{@{}lllllllllll@{}} \toprule 
 & \multicolumn{2}{l}{MobileNetV3} & \multicolumn{2}{l}{DenseNet121} & \multicolumn{2}{l}{ResNet18} & \multicolumn{2}{l}{DINOv1} & \multicolumn{2}{l}{DINOv2} \\ \cmidrule(lr){2-3} \cmidrule(lr){4-5} \cmidrule(lr){6-7} \cmidrule(lr){8-9}  \cmidrule(lr){10-11} 
Testing Datasets & HT-CE & LM & HT-CE & LM & HT-CE & LM & HT-CE & LM & HT-CE & LM \\ \midrule
CIFAR-10 & 17.04 & \textbf{30.24} & 24.23 & \textbf{24.68} & 15.64 & \textbf{27.62} & 21.52 & \textbf{34.5} & 15.88 & \textbf{30.14} \\
CIFAR-100 & 1.79 & \textbf{4.06} & 2.9 & \textbf{3.35} & 1.33 & \textbf{3.96} & 2.22 & \textbf{3.58} & 2.5 & \textbf{5.43} \\
ImageNet-100 & 2.54 & \textbf{3.22} & \textbf{3.08} & 2.86 & 2.58 & \textbf{2.88} & 4.06 &\textbf{4.08} & 2.5 & \textbf{3.86} \\
ImageNet-1000 & 0.26 & \textbf{0.3} & \textbf{0.28} & 0.242 & 0.27 & \textbf{0.4} & \textbf{0.39} & 0.25 & 0.27 & \textbf{0.81} \\
CUB-200 & 0.33 & \textbf{1.02} & \textbf{0.67} & 0.52 & 0.38 & \textbf{0.41} & 0.47 & \textbf{0.67} & 0.28 & \textbf{0.57} \\
Flowers-102 & 0.99 & \textbf{1.06} & \textbf{0.75} & 0.56 & 0.67 & \textbf{0.75} & 1.14 & \textbf{1.32} & 0.28 & \textbf{0.62} \\
Herbarium-19 & \textbf{0.19} & 0.11 & 0.075 & \textbf{0.224} & \textbf{0.075} & 0.035 & 0.075 & 0.075 & 0.15 & 0.15 \\
Oxford-IIIT Pets & \textbf{3.6} & 3.43 & \textbf{4.8} & 2.97 & \textbf{4.77} & 2.78 & \textbf{3.46} & 3.11 & 3.68 & \textbf{3.93} \\
Stanford Cars & 0.61 & 0.61 & \textbf{0.86} & 0.37 & 0.49 & \textbf{0.74} & \textbf{0.74} & 0.61 & 0.49 & \textbf{1.11}\\
\midrule
Average  & 2.97 & \textbf{4.49} & \textbf{4.1} & 3.98 & 2.91 & \textbf{4.4} & 3.79 & \textbf{5.36} & 2.89 & \textbf{5.18} \\
\bottomrule
\end{tabular}
\caption{Top-1 test accuracy comparing High Temperature Cross entropy (HT-CE) against Logit Matching (LM) for use in Prompt Guided Knowledge Distillation. These results are from utilising the mapping function $h$, as defined in Section \ref{sec:method}. Overall, LM outperforms HT-CE, in terms of average accuracy, on four of the five tested vision encoders. We \textbf{bold} the best performance for each comparison.}
\label{tab:pgkd_loss1}
\end{table*}

\begin{table*}[]
\centering
\begin{tabular}{@{}lllllllllll@{}} \toprule 
 & \multicolumn{2}{l}{MobileNetV3} & \multicolumn{2}{l}{DenseNet121} & \multicolumn{2}{l}{ResNet18} & \multicolumn{2}{l}{DINOv1} & \multicolumn{2}{l}{DINOv2} \\ \cmidrule(lr){2-3} \cmidrule(lr){4-5} \cmidrule(lr){6-7} \cmidrule(lr){8-9}  \cmidrule(lr){10-11}
Testing Datasets & HT-CE & LM & HT-CE & LM & HT-CE & LM & HT-CE & LM & HT-CE & LM \\ \midrule
CIFAR-10 &9.75 & \textbf{12.59} & \textbf{12.42} & 8.4 & 9.57 & \textbf{13.62} & 15.63 & \textbf{21.17} & 10.81 & \textbf{19.92} \\
CIFAR-100 & 1.72 & \textbf{1.81} & \textbf{1.75} & 1.6 & 1.08 & \textbf{1.19} & 1.5 & \textbf{2.33} & 2.31 & \textbf{2.88} \\
ImageNet-100 & 1.62 & \textbf{2.56} & \textbf{1.26} & 1.2 & 0.5 & \textbf{1.48} & 2.18 & \textbf{3.68} & 0.74 & \textbf{1.96} \\
ImageNet-1000 & 0.22 & \textbf{0.43} & 0.14 & \textbf{0.15} & 0.11 & \textbf{0.24} & 0.16 & \textbf{0.25} & 0.13 & \textbf{0.29} \\
CUB-200 & \textbf{0.69} & 0.43 & 0.604 & \textbf{0.83} & 0.55 & \textbf{0.71} & 0.52 & \textbf{0.54} & \textbf{0.6} & 0.48 \\
Flowers-102 & \textbf{0.62} & 0.602 & \textbf{1.08} & 0.91 & 0.44 & \textbf{0.83} & 0.67 & \textbf{1.07} & 0.34 & \textbf{0.6} \\
Herbarium-19 & 0.19 & 0.19 & 0.112 & \textbf{0.15} & 0.035 & \textbf{0.22} & 0.26 & 0.26 & 0.075 & \textbf{0.15} \\
Oxford-IIIT Pets & \textbf{3.03} & 2.78 & 2.73 & \textbf{3.46} & \textbf{3.4} & 3.05 & 2.81 & \textbf{5.1} & \textbf{6.69} & 6.11 \\
Stanford Cars & 0.24 & \textbf{0.37} & 0.25 & \textbf{0.49} & 0.49 & 0.49 & \textbf{0.49} & 0.25 & 0.37 & 0.37 \\
\midrule
Average & 2.01 & \textbf{2.42} & \textbf{2.26} & 1.91 & 1.79 & \textbf{2.43} & 2.69 & \textbf{3.85} & 2.45 & \textbf{3.64}\\

\bottomrule
\end{tabular}
\caption{Top-1 test accuracy comparing High Temperature Cross entropy (HT-CE) against Logit Matching (LM) for use in Prompt Guided Knowledge Distillation. These results are from utilising the mapping function $h^{-1}$, as defined in Section \ref{sec:method}. Overall, LM outperforms HT-CE, in terms of average accuracy, on four of the five tested vision encoders. We \textbf{bold} the best performance for each comparison.}
\label{tab:pgkd_loss2}
\end{table*}

\subsection{Impact of Loss Functions}
We provide results comparing the top-1 test accuracy, for comparing the use of different combinations of loss functions. These results are presented across Tables \ref{tab:sup_loss1} to \ref{tab:sup_loss4}. All results follow the trends as discussed in Section \ref{sec:loss}, with no note worthy deviations.

\begin{table}
\resizebox{\linewidth}{!}{%
\begin{tabular}{@{}llllllll@{}} \toprule
Testing Dataset             & MSE   & MSE+PG-KD & MSE+CC    & PG-KD & CC    & CC+PG-KD  & All \\ \midrule
CIFAR-10            & 62.24 & 56.27     & 62.36     & 23.21 & 13.31 & 16.7      & 64.51 \\
CIFAR-100           & 29.77 & 28.49     & 30.01     & 3.47  & 1.06  & 1.53      & 30.75\\
ImageNet-100        & 70.28 & 69.72     & 71.76     & 3.16  & 0.7   & 1.58      & 72.7 \\
ImageNet-1000       & 37.48 & 36.99     & 42.73     & 0.35  & 0.064 & 0.114     & 42.24  \\
CUB-200             & 5.8   & 4.76      & 6.99      & 0.38  & 0.45  & 0.19      & 7.21 \\
Flowers-102         & 4.34  & 4.03      & 4.14      & 0.52  & 0.29  & 1.63      & 4.35\\
Herbarium-19        & 0.075 & 0.112     & 0.075     & 0.26  & 0.15  & 0.15     & 0.11 \\
Oxford-IIIT Pets    & 60.7  & 58.63     & 70.91     & 3.35  & 3.6   & 1.58      & 69.55  \\
Stanford Cars       & 1.23  & 1.11      & 1.72      & 0.25  & 1.35  & 0.62      & 1.1 \\ \midrule
Average             & 30.21 & 28.9     & 32.3     & 3.88  & 2.33  & 2.68      & 32.5 \\
\bottomrule
\end{tabular}%
}

\caption{Top-1 test accuracy of each individual loss function from the \textbf{DenseNet-121}~\cite{Huang2016DenselyCC} vision encoder. MSE: Mean Squared Error; PG-KD: Prompt-Guided Knowledge Distillation with the logit matching variant; CC: Cycle-consistency.}

\label{tab:sup_loss1}
\end{table}

\begin{table}
\resizebox{\linewidth}{!}{%
\begin{tabular}{@{}llllllll@{}} \toprule
Dataset             & MSE   & MSE+PG-KD & MSE+CC    & PG-KD & CC    & CC+PG-KD  & All \\ \midrule
CIFAR-10            & 57.15 & 55.64     & 61.26     & 18.05 & 11.64 & 8.93      & 62.63 \\
CIFAR-100           & 24.16 & 23.82     & 26.31     & 1.32  & 1.1   & 1.37      & 26.77\\
ImageNet-100        & 63.72 & 64.9      & 72.0      & 1.34  & 0.8   & 0.84      & 70.78 \\
ImageNet-1000       & 31.55 & 31.34     & 40.3      & 0.18  & 0.13  & 0.098     & 40.58  \\
CUB-200             & 5.28  & 5.16      & 12.93     & 0.57  & 0.6   & 0.59      & 14.68 \\
Flowers-102         & 3.07  & 3.03      & 5.43      & 0.41  & 0.67  & 0.49      & 6.68\\
Herbarium-19        & 0.037 & 0.075     & 0.0       & 0.11  & 0.11  & 0.15      & 0.03 \\
Oxford-IIIT Pets    & 60.7  & 62.77     & 72.61     & 1.17  & 2.97  & 2.32      & 72.6  \\
Stanford Cars       & 1.11  & 1.35      & 0.98      & 0.61  & 0.12  & 1.35      & 1.59 \\ \midrule
Average             & 27.42 & 27.57     & 32.36     & 2.64  & 2.02  & 1.79      & 32.93 \\
\bottomrule
\end{tabular}%
}

\caption{Top-1 test accuracy of each individual loss function from the \textbf{ResNet18}~\cite{He2015DeepRL} vision encoder.}

\label{tab:loss_performance}
\end{table}

\begin{table}
\resizebox{\linewidth}{!}{%
\begin{tabular}{@{}llllllll@{}} \toprule
Dataset             & MSE   & MSE+PG-KD & MSE+CC    & PG-KD & CC    & CC+PG-KD  & All \\ \midrule
CIFAR-10            & 72.8  & 73.74     & 76.81     & 29.27 & 11.25 & 9.17      & 77.52 \\
CIFAR-100           & 40.45 & 39.62     & 40.87     & 3.04  & 0.78  & 1.05      & 42.68\\
ImageNet-100        & 78.14 & 79.9      & 80.86     & 4.22  & 0.96  & 1.1       & 78.62 \\
ImageNet-1000       & 49.8  & 49.45     & 53.06     & 0.31  & 0.064 & 0.064     & 54.26  \\
CUB-200             & 12.29 & 11.6      & 19.56     & 0.4   & 0.66  & 0.48      & 19.95 \\
Flowers-102         & 7.29  & 7.16      & 12.3      & 0.86  & 1.12  & 0.91      & 13.69\\
Herbarium-19        & 0.26  & 0.15      & 0.15      & 0.15  & 0.037 & 0.075     & 0.187 \\
Oxford-IIIT Pets    & 74.63 & 75.25     & 77.39     & 3.98  & 1.88  & 6.19      & 77.84  \\
Stanford Cars       & 1.35  & 1.6       & 2.21      & 0.37  & 0.74  & 0.61      & 1.47 \\ \midrule
Average             & 37.45 & 37.61     & 40.35     & 4.73  & 1.94  & 2.18      & 40.69 \\
\bottomrule
\end{tabular}%
}

\caption{Top-1 test accuracy of each individual loss function from the \textbf{DinoV1}~\cite{dino2021} vision encoder.}

\label{tab:loss_performance}
\end{table}

\begin{table}
\resizebox{\linewidth}{!}{%
\begin{tabular}{@{}llllllll@{}} \toprule
Dataset             & MSE   & MSE+PG-KD & MSE+CC    & PG-KD & CC    & CC+PG-KD  & All \\ \midrule
CIFAR-10            & 94.07 & 93.63     & 93.45     & 17.84 & 8.76  & 10.3      & 94.35 \\
CIFAR-100           & 63.34 & 62.87     & 65.35     & 2.59  & 0.65  & 0.5      & 65.43\\
ImageNet-100        & 84.3  & 85.76     & 86.96     & 2.22  & 0.86  & 0.66       & 86.96 \\
ImageNet-1000       & 58.51 & 58.5      & 63.33     & 0.28  & 0.078 & 0.09     & 63.62  \\
CUB-200             & 18.42 & 19.42     & 28.87     & 0.79  & 0.56  & 0.74      & 29.32 \\
Flowers-102         & 18.78 & 15.79     & 22.75     & 1.14  & 0.68  & 0.63      & 25.32\\
Herbarium-19        & 0.15  & 0.19      & 0.11      & 0.04  & 0.075 & 0.15     & 0.187 \\
Oxford-IIIT Pets    & 80.98 & 81.3      & 83.13     & 4.14  & 1.85  & 3.79      & 86.96  \\
Stanford Cars       & 5.41  & 4.91      & 5.63      & 0.25  & 0.98  & 0.86      & 5.28 \\ \midrule
Average             & 47.11 & 46.93     & 49.95     & 3.25  & 1.61  & 1.97    & 50.83 \\
\bottomrule
\end{tabular}%
}

\caption{Top-1 test accuracy of each individual loss function from the \textbf{DinoV2}~\cite{dinov22023} vision encoder.}

\label{tab:sup_loss4}
\end{table}

\subsection{Size of Training Dataset}
We provide results comparing the top-1 test accuracy of our Zoom-shot method against Linear Aligner (LA) \cite{t2c2023} when using different amounts of training data. These results are presented across Tables \ref{tab:supp_lim_ours1} to \ref{tab:sup_lim_ours4}, with Table \ref{tab:supp_lim_ours1} (Zoom-shot) compared against Table \ref{tab:sup_lim_la1} (LA), Table \ref{tab:sup_lim_ours2} (Zoom-shot) compared against Table \ref{tab:sup_lim_la2} (LA) and so on. Same as before, all results follow the trends as discussed in Section \ref{sec:ab_data}, with no note worthy deviations.

\begin{table}
\centering
\resizebox{0.8\linewidth}{!}{%
\begin{tabular}{@{}llllll@{}} \toprule
Dataset         & 100\% & 20\%  & 5\%   & 1\%   & 1\% (20 epochs) \\ \midrule
CIFAR-10        & 60.85 & 64.51 & 63.6 & 40.07  & 62.74 \\
CIFAR-100       & 30.36 & 30.75 & 28.39  & 18.24  & 29.98 \\
ImageNet-100    & 72.32 & 72.7  & 73.02  & 43.86  & 74.16 \\
ImageNet-1000   & 40.64 & 41.24 & 40.09  & 14.39  & 41.37 \\
CUB-200         & 5.37  & 7.21  & 6.94  & 1.14  & 7.25 \\
Flowers-102     & 2.65  & 4.35  & 2.13  & 1.14  & 3.38 \\
Herbarium-19    & 0.075   & 0.11  & 0.075 & 0.15  & 0.04 \\
Oxford-IIIT Pets& 68.74 & 69.55 & 72.58  & 39.89  & 69.47 \\
Stanford Cars   & 1.35   & 1.1   & 1.11  & 0.61  & 2.1 \\ \midrule
Average         & 31.37 & 32.5 & 31.99  & 17.72  & 32.28 \\
\bottomrule
\end{tabular}%
}

\caption{Top-1 test accuracy of the \textbf{DenseNet-121}~\cite{Huang2016DenselyCC} vision encoder using Zoom-shot on different amounts of ImageNet~\cite{deng_imagenet_2009} training data. In the final column, we show that simply extending the training time to 20 epochs allows our method to nearly match the performance at 20\% training data. All other columns are trained for 1 epoch. These results are compared against Table \ref{tab:sup_lim_la1}.
}

\label{tab:supp_lim_ours1}
\end{table}

\begin{table}
\resizebox{\linewidth}{!}{%
\begin{tabular}{@{}lllllll@{}} \toprule
Dataset & \makecell{100\%\\(6 epochs)}&\makecell{20\%\\(6 epochs)}&\makecell{1\%\\(6 epochs)}&\makecell{1\%\\(20 epochs)}& \makecell{1\%\\(120 epochs)} \\ \midrule
CIFAR-10            & 58.79  & 64.46 & 53.72 & 55.33 & 57.77 \\
CIFAR-100           & 27.94  & 28.25 & 22.72 & 26.88 & 27.79 \\
ImageNet-100        & 70.48  & 70.14 & 58.86 & 68.62 & 71.02 \\
ImageNet-1000       & 37.29  & 37.43 & 25.85 & 34.49 & 36.97 \\
CUB-200             & 4.13  & 5.5   & 2.18  & 3.64  & 4.38  \\
Flowers-102         & 2.12  & 4.14  & 1.67  & 2.05  & 2.34  \\
Herbarium-19        & 0.075  & 0.11  & 0.0   & 0.04  & 0.075 \\
Oxford-IIIT Pets    & 60.92  & 61.57 & 50.78 & 59.39 & 62.39  \\
Stanford Cars       & 0.98  & 1.1   & 1.11  & 0.98  & 1.11  \\ \midrule
Average             & 29.19  & 30.3 & 24.1  & 27.94  & 29.32 \\
\bottomrule
\end{tabular}%
}

\caption{Top-1 test accuracy of the \textbf{DenseNet-121}~\cite{Huang2016DenselyCC}  vision encoder using Linear Aligner~\cite{t2c2023} on different amounts of ImageNet~\cite{deng_imagenet_2009} training data. In the final column, we show that simply extending the training time to 120 epochs allows the method to nearly match the performance at 20\% training data. These results are compared against Table \ref{tab:supp_lim_ours1}.}

\label{tab:sup_lim_la1}
\end{table}

\begin{table}
\centering
\resizebox{0.8\linewidth}{!}{%
\begin{tabular}{@{}llllll@{}} \toprule
Dataset         & 100\% & 20\%  & 5\%   & 1\%   & 1\% (20 epochs) \\ \midrule
CIFAR-10        & 58.14 & 62.63 & 62.09 & 42.09  & 59.17 \\
CIFAR-100       & 25.99 & 26.77 & 23.45  & 12.36  & 25.39 \\
ImageNet-100    & 70.7 & 70.78  & 66.86 & 32.12  & 69.92 \\
ImageNet-1000   & 41.14 & 40.58 & 35.65  & 8.71  & 38.74 \\
CUB-200         & 11.86  & 14.68  & 7.58  & 0.69  & 7.94 \\
Flowers-102     & 3.89  & 6.68  & 3.17  & 1.4  & 3.22 \\
Herbarium-19    & 0.0   & 0.03  & 0.075 & 0.15  & 0.04 \\
Oxford-IIIT Pets& 72.58 & 72.6 & 71.79  & 33.77  & 71.6 \\
Stanford Cars   & 1.84  & 1.59   & 1.72  & 0.5  & 1.35 \\ \midrule
Average         & 31.79 & 32.93 & 30.27  & 14.64  & 30.82 \\
\bottomrule
\end{tabular}%
}

\caption{Top-1 test accuracy using the \textbf{ResNet-18}~\cite{He2015DeepRL} vision encoder using Zoom-shot on different amounts of ImageNet~\cite{deng_imagenet_2009} training data. These results are compared against Table \ref{tab:sup_lim_la2}.
}

\label{tab:sup_lim_ours2}
\end{table}

\begin{table}
\resizebox{\linewidth}{!}{%
\begin{tabular}{@{}lllllll@{}} \toprule
Dataset & \makecell{100\%\\(6 epochs)}&\makecell{20\%\\(6 epochs)}&\makecell{1\%\\(6 epochs)}&\makecell{1\%\\(20 epochs)}& \makecell{1\%\\(120 epochs)} \\ \midrule
CIFAR-10            & 52.67  & 55.12 & 47.84 & 47.39 & 53.63 \\
CIFAR-100           & 22.31  & 23.38 & 15.42 & 19.08 & 21.78 \\
ImageNet-100        & 63.4  & 63.36 & 49.6  & 60.78 & 61.7 \\
ImageNet-1000       & 31.07  & 31.28 & 17.78 & 27.22 & 30.47 \\
CUB-200             & 3.66  & 5.4   & 1.76  & 3.21  & 3.69  \\
Flowers-102         & 2.1  & 2.74  & 1.69  & 2.29  & 2.44  \\
Herbarium-19        & 0.11  & 0.11  & 0.11  & 0.04  & 0.075 \\
Oxford-IIIT Pets    & 60.78  & 62.36 & 46.58 & 58.52 & 60.64  \\
Stanford Cars       & 0.99  & 1.72  & 0.74  & 1.11  & 0.86  \\ \midrule
Average             & 26.34  & 27.27 & 20.17 & 24.4  & 26.14 \\
\bottomrule
\end{tabular}%
}

\caption{Top-1 test accuracy of the \textbf{ResNet-18}~\cite{He2015DeepRL} vision encoder using Linear Aligner~\cite{t2c2023} on different amounts of ImageNet~\cite{deng_imagenet_2009} training data. These results are compared against Table \ref{tab:sup_lim_ours2}.}

\label{tab:sup_lim_la2}
\end{table}

\begin{table}
\centering
\resizebox{0.8\linewidth}{!}{%
\begin{tabular}{@{}llllll@{}} \toprule
Dataset         & 100\% & 20\%  & 5\%   & 1\%   & 1\% (20 epochs) \\ \midrule
CIFAR-10        & 75.39 & 77.52 & 76.66 & 71.67  & 77.06 \\
CIFAR-100       & 43.35 & 42.68 & 41.18  & 36.3  & 40.34 \\
ImageNet-100    & 81.18 & 78.62  & 80.0 & 72.52  & 80.7 \\
ImageNet-1000   & 55.01 & 54.26 & 53.14  & 40.42  & 53.25 \\
CUB-200         & 18.9  & 19.95 & 17.19  & 6.8  & 16.9 \\
Flowers-102     & 10.02  & 13.69  & 8.03  & 3.9  & 6.96 \\
Herbarium-19    &  0.299  & 0.187  & 0.15 & 0.5  & 0.11 \\
Oxford-IIIT Pets&  77.61 & 77.84 & 78.88  & 68.7  & 78.4 \\
Stanford Cars   & 1.23  & 1.47   & 1.97  & 0.5  & 1.6 \\ \midrule
Average         & 40.33 & 40.69 & 39.69  & 33.48  & 39.48 \\
\bottomrule
\end{tabular}%
}

\caption{Top-1 test accuracy of the \textbf{DINOv1}~\cite{dino2021} vision encoder using Zoom-shot on different amounts of ImageNet~\cite{deng_imagenet_2009} training data. These results are compared against Table \ref{tab:sup_lim_la3}.
}

\label{tab:sup_lim_ours3}
\end{table}

\begin{table}
\resizebox{\linewidth}{!}{%
\begin{tabular}{@{}lllllll@{}} \toprule
Dataset & \makecell{100\%\\(6 epochs)}&\makecell{20\%\\(6 epochs)}&\makecell{1\%\\(6 epochs)}&\makecell{1\%\\(20 epochs)}& \makecell{1\%\\(120 epochs)} \\ \midrule
CIFAR-10            & 70.13  & 72.97 & 66.2  & 71.17 & 70.93 \\
CIFAR-100           & 40.03  & 41.18 & 32.96 & 36.41 & 40.0 \\
ImageNet-100        & 79.2  & 77.86 & 65.52 & 76.16 & 78.52 \\
ImageNet-1000       & 50.52  & 49.02 & 31.18 & 44.76 & 48.6 \\
CUB-200             & 9.01  & 11.94 & 2.43  & 6.7   & 8.34  \\
Flowers-102         & 4.52  & 6.42  & 3.24  & 3.38  & 4.62  \\
Herbarium-19        & 0.26  & 0.187 & 0.04  & 0.34  & 0.11 \\
Oxford-IIIT Pets    & 75.17  & 74.62 & 64.68 & 72.45 & 74.03  \\
Stanford Cars       & 0.86  & 1.47  & 0.49  & 0.86  & 0.86  \\ \midrule
Average             & 36.63  & 37.3  & 29.64 & 34.69 & 36.22 \\
\bottomrule
\end{tabular}%
}

\caption{Top-1 test accuracy of the \textbf{DINOv1}~\cite{dino2021} vision encoder using Linear Aligner~\cite{t2c2023} on different amounts of ImageNet~\cite{deng_imagenet_2009} training data. These results are compared against Table \ref{tab:sup_lim_ours3}.}

\label{tab:sup_lim_la3}
\end{table}

\begin{table}[t]
\centering
\resizebox{0.8\linewidth}{!}{%
\begin{tabular}{@{}llllll@{}} \toprule
Dataset         & 100\% & 20\%  & 5\%   & 1\%   & 1\% (20 epochs) \\ \midrule
CIFAR-10        & 94.63 & 94.35 & 94.81 & 91.27  & 94.71 \\
CIFAR-100       & 66.87 & 65.42 & 66.79  & 54.53  & 64.48 \\
ImageNet-100    & 88.54 & 86.96  & 85.94 & 78.98  & 85.56 \\
ImageNet-1000   & 64.1 & 63.62 & 62.8  & 48.88  & 62.71\\
CUB-200         & 27.46  & 29.32 & 28.27  & 17.36  & 26.06 \\
Flowers-102     & 19.43  & 25.32  & 27.57  & 6.29  & 20.54 \\
Herbarium-19    & 0.15   & 0.187  & 0.075 & 0.15  & 0.11 \\
Oxford-IIIT Pets& 83.78 & 86.96 & 53.53  & 74.7  & 84.25 \\
Stanford Cars   & 5.16  & 5.28   & 6.27  & 2.95  & 5.41 \\ \midrule
Average         & 50.01 & 50.82 & 47.34  & 41.65  & 49.31 \\
\bottomrule
\end{tabular}%
}

\caption{Top-1 test accuracy of the \textbf{DINOv2}~\cite{dinov22023} vision encoder using Zoom-shot on different amounts of ImageNet~\cite{deng_imagenet_2009} training data. These results are compared against Table \ref{tab:sup_lim_la4}. 
}

\label{tab:sup_lim_ours4}
\end{table}

\begin{table}[t]
\resizebox{\linewidth}{!}{%
\begin{tabular}{@{}lllllll@{}} \toprule
Dataset & \makecell{100\%\\(6 epochs)}&\makecell{20\%\\(6 epochs)}&\makecell{1\%\\(6 epochs)}&\makecell{1\%\\(20 epochs)}& \makecell{1\%\\(120 epochs)} \\ \midrule
CIFAR-10            & 94.09  & 93.8  & 83.64 & 93.61 & 94.15 \\
CIFAR-100           & 65.41  & 64.47 & 47.86 & 60.94 & 62.77 \\
ImageNet-100        & 87.02  & 85.34 & 66.64 & 84.52 & 84.6 \\
ImageNet-1000       & 60.36  & 59.67 & 36.03 & 56.21 & 59.51 \\
CUB-200             & 16.71  & 18.74 & 4.76  & 14.81 & 18.38  \\
Flowers-102         & 14.13  & 16.55 & 3.69  & 14.2  & 18.41  \\
Herbarium-19        & 0.3  & 0.037 & 0.15  & 0.04  & 0.15 \\
Oxford-IIIT Pets    & 80.43   & 80.05 & 58.76 & 79.7  & 80.24  \\
Stanford Cars       & 3.93  & 3.31  & 2.33  & 4.91  & 4.42  \\ \midrule
Average             &  46.93 & 46.89 & 33.76 & 45.44 & 46.96 \\
\bottomrule
\end{tabular}%
}

\caption{Top-1 test accuracy of the \textbf{DINOv2}~\cite{dinov22023} vision encoder using Linear Aligner~\cite{t2c2023} on different amounts of ImageNet~\cite{deng_imagenet_2009} training data. These results are compared against Table \ref{tab:sup_lim_ours4}.}

\label{tab:sup_lim_la4}
\end{table}


\end{document}